\documentclass[10pt,twocolumn,letterpaper]{article}

\usepackage{cvpr}      % To produce the REVIEW version
\usepackage[accsupp]{axessibility} % Improves PDF readability for those with disabilities.
\usepackage{algorithmic}
\usepackage{algorithm}
\usepackage{amsmath} 
\usepackage{multirow} 
\usepackage{makecell}
\usepackage{diagbox}
\usepackage{bm}

\usepackage[dvipsnames]{xcolor}

\definecolor{cvprblue}{rgb}{0.21,0.49,0.74}
\usepackage[pagebackref,breaklinks,colorlinks,citecolor=cvprblue]{hyperref}

\def\paperID{15060} % *** Enter the Paper ID here
\def\confName{CVPR}
\def\confYear{2024}

\title{Generalizable Face Landmarking Guided by Conditional Face Warping}

\author{{Jiayi Liang$^{1}$\quad Haotian Liu$^{1}$\thanks{The first two authors contributed equally to this work.}\quad Hongteng Xu$^{2,3}$\quad Dixin Luo$^{1,4}$\thanks{Corresponding author.}}\\
$^1$School of Computer Science and Technology, 
Beijing Institute of Technology, Beijing\\
$^2$Gaoling School of Artificial Intelligence, 
Renmin University of China, Beijing\\
$^3$Beijing Key Laboratory of Big Data Management and Analysis Methods, Beijing\\
$^4$Key Laboratory of Artificial Intelligence,  Ministry of Education, Shanghai\\
{\tt\small \{jiayi.liang, haotianliu, dixin.luo\}@bit.edu.cn, hongtengxu@ruc.edu.cn}
}

\begin{document}
\maketitle

\begin{abstract}
    As a significant step for human face modeling, editing, and generation, face landmarking aims at extracting facial keypoints from images. 
    A generalizable face landmarker is required in practice because real-world facial images, e.g., the avatars in animations and games, are often stylized in various ways. 
    However, achieving generalizable face landmarking is challenging due to the diversity of facial styles and the scarcity of labeled stylized faces. 
    In this study, we propose a simple but effective paradigm to learn a generalizable face landmarker based on labeled real human faces and unlabeled stylized faces.
    Our method learns the face landmarker as the key module of a conditional face warper. 
    Given a pair of real and stylized facial images, the conditional face warper predicts a warping field from the real face to the stylized one, in which the face landmarker predicts the ending points of the warping field and provides us with high-quality pseudo landmarks for the corresponding stylized facial images. 
    Applying an alternating optimization strategy, we learn the face landmarker to minimize $i)$ the discrepancy between the stylized faces and the warped real ones and $ii)$ the prediction errors of both real and pseudo landmarks. 
    Experiments on various datasets show that our method outperforms existing state-of-the-art domain adaptation methods in face landmarking tasks, leading to a face landmarker with better generalizability. 
    Code is available at \href{https://plustwo0.github.io/project-face-landmarker}{https://plustwo0.github.io/project-face-landmarker}.
\end{abstract}

\section{Introduction}\label{sec1}

Face landmarking seeks to extract human facial keypoints (\eg, eyes, nose, facial contour, and so on) from facial images. 
This task is important for many applications in the field of computer vision and graphics, such as face recognition~\citep{taigman2014deepface,masi2016pose,liu2017sphereface,yang2017neural,agbolade2020landmark}, face stylization~\citep{cao2018carigans}, and 3D face reconstruction~\citep{dou2017end,roth2015unconstrained,liu2016joint}. 
Currently, many open-source and commercial face landmarkers~\citep{Zhou_2013_ICCV_Workshops,kazemi2014one,wu2018look} have been developed and achieved encouraging performance in this task. 

\begin{figure}[t]
    \centering
    \includegraphics[width=0.44\textwidth]{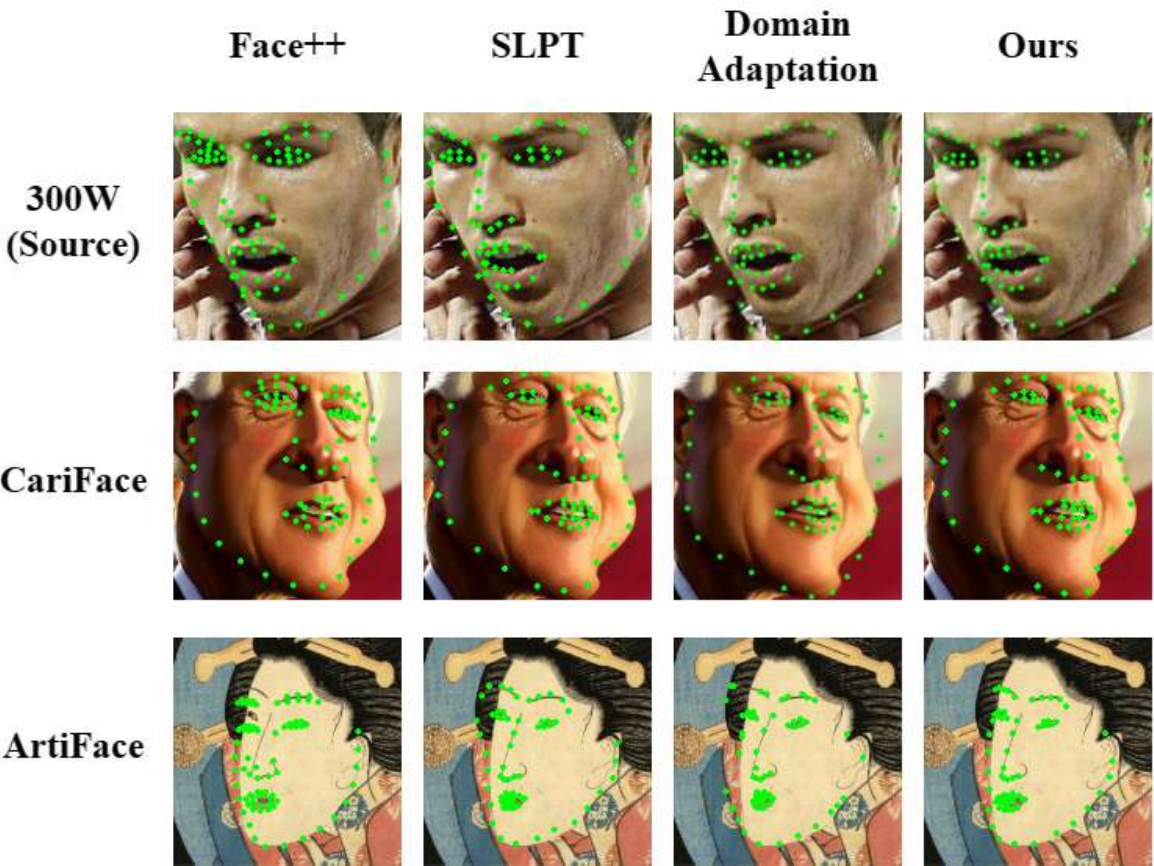}
    \caption{
    Both commercial software like Face++ and open-source method like SLPT~\cite{xia2022sparse} work well on landmarking real faces (e.g., those in 300W~\citep{Sagonas_Tzimiropoulos_Zafeiriou_Pantic_2014}) while achieving suboptimal performance when landmarking stylized faces (e.g., those in CariFace~\citep{Cai_Guo_Peng_Zhang_2020} and ArtiFace~\citep{yaniv2019face}). 
    While existing domain adaptation method does not improve the performance significantly, our method achieves a generalizable face landmarker for various facial images.
    }
    \label{fig:failure_cases}
\end{figure}

Most existing face landmarkers are designed and trained for landmarking real human faces, while the rapid development of AIGC applications, such as artistic character creation and cartoon generation~\citep{gong2020autotoon, Aliakbarian_2022_CVPR}, leads to a massive increase in demands for landmarking stylized facial images.
Unfortunately, as shown in~\cref{fig:failure_cases}, existing face landmarkers often fail to landmark stylized facial images. 
Even if applying state-of-the-art domain adaptation strategies~\citep{ganin2015unsupervised, li2019bidirectional,Zhu_2017_ICCV,zhu2023representation,liu2023learnable}, the generalizability of the learned landmarkers in the stylized facial image domain is still unsatisfactory. 
Essentially, traditional face landmarkers work well on real human faces because of the relatively stable geometry of real human faces and the sufficient labeled facial images. 
These two conditions, however, become questionable when landmarking stylized faces --- the stylized facial images often have various facial styles, and manually landmarking such stylized faces is much more time-consuming than landmarking real faces. 
As a result, learning a generalizable face landmarker becomes a challenging task.

\begin{figure}[t]
    \centering
    \includegraphics[height=6.8cm]{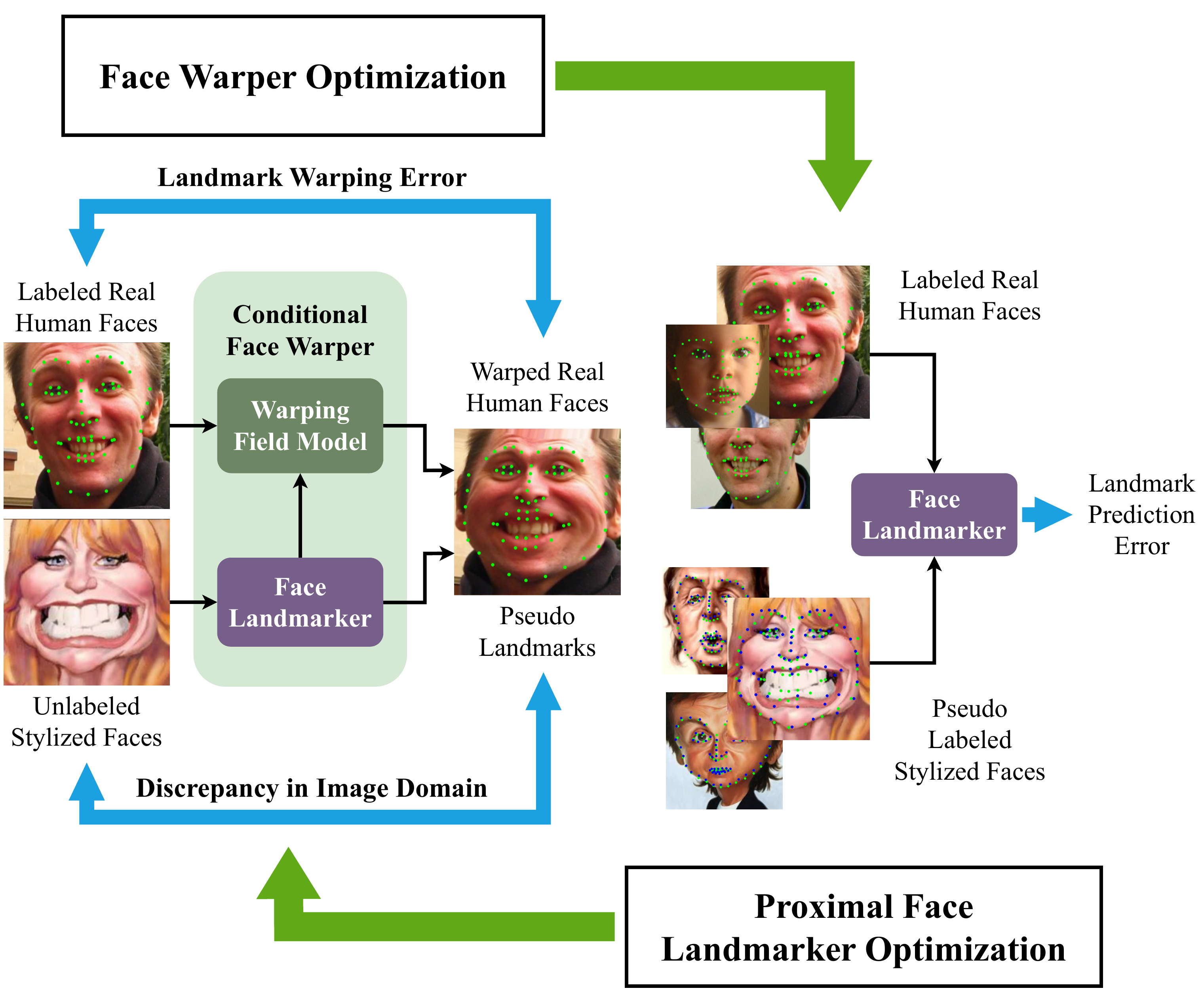}
    \caption{The scheme of our proposed method for learning a generalizable face landmarker. 
    }
    \label{fig:overview}
\end{figure}

In this paper, we propose a simple but effective paradigm for learning a generalizable face landmarker, overcoming the challenges caused by the diversity of facial styles and the scarcity of labeled stylized faces. 
As illustrated in~\cref{fig:overview}, given labeled real human facial images and unlabeled stylized facial images, we learn a face landmarker embedded in a conditional face warper. 
The face warper aims to deform real human faces according to stylized facial images, generating warped faces and corresponding warping fields. 
The face landmarker, as the key module of the warper, predicts the ending points of the warping fields and thus provides us with pseudo landmarks for the stylized facial images. 
The warping field is parametrized by a polyharmonic interpolation model.
Under the guidance of the conditional face warping, we learn the face landmarker in an alternating optimization framework: The face landmarker is updated to $i)$ minimize the discrepancy between the stylized faces and the corresponding warped real human faces and $ii)$ minimize the prediction errors of both real and pseudo landmarks. 
In the first step, the face landmarker is learned associated with the warping field model, while in the second step, the face landmarker is updated with proximal regularization.

The extensive experiments in various face landmarking tasks demonstrate the effectiveness of learning method. 
The impacts of different loss functions and data settings on our learning method are analyzed through detailed ablation studies. 
Experimental results show that our learning method results in a face landmarker that is generalizable to the facial images with different styles, which outperforms representative domain adaptation methods consistently in various stylized face landmarking tasks.

\section{Related Work}
\subsection{Face Landmarking}
Given facial images, the early face landmarking methods learn regression models to predict landmark coordinates directly~\citep{cao2014face,lv2017deep}. 
The models are often parameterized by neural networks like Transformer~\citep{vaswani2017attention}, capturing facial attributes that have been proven to be crucial for landmark prediction~\cite{Li_2022_CVPR}. 
The coordinate regression is achieved in a coarse-to-fine framework~\citep{Zhou_2013_ICCV_Workshops}, leading to a cascading landmarking pipeline.
Recently, to make the landmark prediction robust to the variations in pose, scale, and occlusion, DAN~\citep{Kowalski_2017_CVPR_Workshops} introduces the novel use of heatmaps and extracts features from the entire face rather than local patches around landmarks. 
SBR~\citep{dong2018supervision} utilizes the registration of synthesized images to provide supervisory signals for training.
Adaptive Wing Loss~\citep{Wang_2019_ICCV} is proposed to address the imbalance between foreground pixels and background pixels by analysis of the main drawbacks of different loss functions. 
HRNet~\citep{wang2020deep} produces high-resolution maps by connecting and exchanging information via merging multi-scale picture features across many branches. 

As aforementioned, most existing face landmarkers are learned for real human faces, which can not be adapted directly to stylized faces (e.g., cartoon and artistic faces). 
Essentially, treating labeled real human faces as a source domain and unlabeled stylized faces as a target domain, we can learn a generalizable face landmark by solving a domain adaptation (DA) problem~\citep{long2015learning,long2016unsupervised}.
Accordingly, many DA techniques~\cite{liu2023learnable,wu2023domain} have potential in our problem, including the classic metric learning-based methods (e.g., CORAL~\citep{sun2017correlation}, contrastive domain discrepancy~\citep{kang2019contrastive}, and maximum mean discrepancy~\citep{gretton2006kernel}) and the recent adversarial learning-based methods~\citep{ganin2015unsupervised, Zhu_2017_ICCV, pei2018multi,zhang2018importance}.
However, in the following content, we will show that directly applying these DA techniques often fails to achieve generalizable face landmarking because of the significant gap between the source and target face domains.

\subsection{Face Warping}
Face warping is a technique that involves geometrically deforming source facial images to specified target shapes.
The key step of this task is predicting a warping field between the source and target images that captures the shifts of image pixels.
To achieve this aim, DST~\citep{Kim_Kolkin_Salavon_Shakhnarovich_2020} and FoA~\citep{yaniv2019face} find matching keypoints between source and target images and then generate a dense warping field through data interpolation~\citep{cole2017synthesizing}.
Instead of matching keypoints, some methods learn neural networks to predict dense warping fields directly based on paired images, e.g., Flownet~\citep{dosovitskiy2015flownet}, AutoToon~\citep{gong2020autotoon}, RAFT~\citep{teed2020raft}, and their variants~\citep{liu2021learning}. 
However, these methods require the paired images to be similar to each other, which cannot capture significant deformations between the faces in different domains. 

Compared to face stylization~\citep{gatys2015neural,selim2016painting,shi2019warpgan,karras2019style}, face warping is a relatively easier task because it only considers the deformation of shapes while ignoring the transfer of textures. 
However, it should be noted that this task is more relevant to face landmarking, in which the warping field provides us with strong evidence to shift face landmarks of source faces to target ones~\cite{wu2019landmark}. 
Inspired by such a strong correlation, we develop the proposed learning paradigm.

% Face cartoonization aims to generate an alike cartoon face for any person automatically. 
% The typical way of face cartoonization is approached by synthesizing style texture based on the semantic content of input image \citep{selim2016painting} inspired by the success of neural style transfer \citep{gatys2015neural}. 
% Neural style transfer \citep{gatys2015neural} extracts style from the shallow layers of a CNN and further extracts contents from the deeper layers. 
% The effectiveness of earlier style transfer techniques, nevertheless, is still restricted to color and texture and does not succeed in transferring geometric style. 
% By training on large-scale datasets, GANs~\citep{goodfellow2020generative} can generate high-quality cartoon-like representations with better preservation of facial details and artistic styles. 
% These works~\citep{shi2019warpgan,wu2019landmark,karras2019style,karras2020analyzing,yang2022pastiche} achieve impressive results compared to the texture-only style transfer methods. 

\section{Proposed Method}
Denote $\mathcal{X}$ as the image space and $\mathcal{Y}$ as the landmark space, respectively.
In this work, we observe a set of labeled real human faces, i.e., $\mathcal{D}^{(L)}=\{\bm{X}_i^{(L)}, \bm{Y}_i^{(L)}\}_{i=1}^{N_L}\subset  \mathcal{X}_R\times \mathcal{Y}$ and a set of unlabeled stylized faces, i.e., $\mathcal{D}^{(U)}=\{\bm{X}_i^{(U)}\}_{i=1}^{N_U}\subset \mathcal{X}_S$, where $\mathcal{X}_R,\mathcal{X}_S\subset\mathcal{X}$ correspond to the real and stylized face domains, respectively.
Each $\bm{X}_i\in\mathbb{R}^{H\times W \times 3}$ represents an image, and each $\bm{Y}_i=[\bm{y}_{i,k}]\in\mathbb{R}^{2\times K}$ records $K$ face landmark coordinates, where $\bm{y}_{i,k}\in\mathbb{R}^2$. 
We aim to learn a face landmarker, denoted as $f_{\theta}:\mathcal{X}\mapsto\mathcal{Y}$, where $\theta$ is the model parameter. 
The model should be able to predict face landmarks from facial images and moreover, generalize to both $\mathcal{X}_R$ and $\mathcal{X}_S$. 
To achieve this aim, we embed the face landmarker into a conditional face warper and learn it associated with a parametric warping field predictor in an alternating optimization framework, as illustrated in~\cref{fig:overview}.

\subsection{Face Landmarking Guided by Face Warping} 
In this study, we take the SLPT model~\citep{xia2022sparse} as the backbone of our face landmarker. 
Given a stylized face $\bm{X}_i^{(U)}$, the face landmarker predicts its landmarks as $\widehat{\bm{Y}}_i^{(U)}=[\hat{\bm{y}}_{i,k}^{(U)}]=f_{\theta}(\bm{X}_i^{(U)})$. 
At the same time, we can sample a labeled real face $(\bm{X}_j^{(L)},\bm{Y}_{j}^{(L)})\sim\mathcal{D}^{L}$.
Treating the labeled and predicted landmarks as keypoints, we can model a warping field from the real face to the stylized one by the following polyharmonic interpolation model~\cite{Glasbey_Mardia_2002}: 
\begin{eqnarray}\label{eq:semi}
\begin{aligned}
    w_{i,\gamma}(\bm{y}) = \sideset{}{_{k=1}^{K}}\sum\bm{\omega}_k\phi(\|\bm{y}-\hat{\bm{y}}_{i,k}^{(U)}\|_2) + \bm{V}\bm{y} + \bm{b},
\end{aligned}
\end{eqnarray}
where $\gamma=\{\{\bm{\omega}_k\in\mathbb{R}^{2}\}_{k=1}^{K},\bm{V}\in\mathbb{R}^{2\times 2},\bm{b}\in\mathbb{R}^2\}$ correspond to the parameters of the warping field.
As shown in~\eqref{eq:semi}, the vector $\bm{y}$ denotes the u-v coordinate of a pixel in the stylized facial image, and $w_{i,\gamma}(\bm{y})$ gives the inverse mapping from the pixel $\bm{y}$ to a coordinate in the real human facial image, conditioned on $\bm{X}_i^{(U)}$.
The first term $\sum_{k=1}^{K}\omega_k\phi(\|\bm{y}-\hat{\bm{y}}_{i,k}^{(U)}\|_2)$ achieves nonparametric regression for modeling nonrigid deformations, in which $\phi(r)$ is a predefined thin-plate spline function. 
The second term $\bm{V}\bm{y} + \bm{b}$ is a linear parametric model capturing the rigid transformation of $\bm{y}$.

For each pixel coordinate $\bm{y}\in\{1,...,H\}\times\{1,...,W\}$, we can trace it back to the real human facial image based on $w_{i,\gamma}(\bm{y})$ and obtain the pixel color as $\bm{X}_j^{(L)}(w_{i,\gamma}(\bm{y}))$. 
Accordingly, with the grid sampler constructed via inverse mapping function $w_{i,\gamma}$, we obtain the warped real human facial image conditioned on $\bm{X}_i^{(U)}$, denoted as $\widehat{\bm{X}}_{j|i}^{(L)}$.
For $\bm{y}\in\{1,...,H\}\times\{1,...,W\}$ and $j=1,...,N_L$, we have 
\begin{eqnarray}\label{eq:warp}
\begin{aligned}
    \widehat{\bm{X}}_{j|i}^{(L)}(\bm{y}) = \bm{X}_j^{(L)}(w_{i,\gamma}(\bm{y})).
\end{aligned}
\end{eqnarray}
Unlike WarpGAN~\citep{shi2019warpgan}, which generates the warped face by predicting dense keypoints and their displacements by two fully-connected layers during training, we directly use the predicted and observed landmarks to define sparse displacements and estimate other pixels' displacements by spline-based interpolation, which improves computational efficiency significantly. 
Moreover, by applying the warping field model with limited degree-of-freedom (i.e., few learnable parameters), we can focus more on the learning of the face landmarker in the training phase.

Specifically, the warped face together with the warping field provides a useful guidance for the learning of the face landmarker.
In particular, we formulate the learning problem of the face landmarker as follows:
\begin{eqnarray}\label{eq:opt}
\begin{aligned}
    &\sideset{}{_{\theta,\gamma}}\min \underbrace{\sideset{}{_{j=1}^{N_L}}\sum\|f_{\theta}(\bm{X}_j^{(L)})-\bm{Y}_j^{(L)}\|_F^2}_{\text{Landmarking error in the source domain}}  \\
    &\quad+\underbrace{\sideset{}{_{i=1}^{N_U}}\sum\sideset{}{_{j=1}^{N_L}}\sum\|\nabla\widehat{\bm{X}}_{j|i}^{(L)}-\nabla\bm{X}_i^{(U)}\|_F^2}_{\text{Discrepancy of image gradient}} \\
    &\quad+\underbrace{\sideset{}{_{i=1}^{N_U}}\sum\sideset{}{_{j=1}^{N_L}}\sum\|w_{i,\gamma}(\widehat{\bm{Y}}_i^{(U)}) - \bm{Y}_j^{(L)}\|_F^2}_{\text{Landmark warping error}},
\end{aligned}
\end{eqnarray}
where $\|\cdot\|_F$ represents the Frobenius norm of matrix.
In~\eqref{eq:opt}, the first term is the landmarking error for real faces, which corresponds to the data fidelity loss in the source domain.
The second term measures the discrepancy between the stylized face and the warped real face in the gradient field, in which the gradient operation $\nabla$ is implemented by the Sobel operator.
The third term is the landmark warping error. 
Both the second and third terms are determined jointly by the landmarker $f_{\theta}$ and warping field model $w_{i,\gamma}$.

\subsection{Alternating Optimization Strategy}\label{sec: strategy}
The optimization problem in~\eqref{eq:opt} is non-convex because the landmarker is implemented by a neural network and is coupled to the warping field model.
As a result, learning $\theta$ and $\gamma$ jointly often falls into an undesired local optimum even an unstable saddle point.
To mitigate this issue, we propose an alternating optimization framework.
In principle, we can decompose the optimization problem in~\eqref{eq:opt} into the following two subproblems and solve them iteratively.

\begin{itemize}
\item \textbf{Face Warper Optimization:} 
The first subproblem corresponds to the optimization of the face warper, i.e.,
\begin{eqnarray}\label{eq:opt_warper}
    \begin{aligned}
    \theta^{(1)},\gamma^{(1)}=&\arg\sideset{}{_{\theta,\gamma}}\min \sideset{}{_{i,j}}\sum\|\nabla\widehat{\bm{X}}_{j|i}^{(L)}-\nabla\bm{X}_i^{(U)}\|_F^2 \\
    &\quad+\sideset{}{_{i,j}}\sum\|w_{i,\gamma}(\widehat{\bm{Y}}_i^{(U)}) - \bm{Y}_j^{(L)}\|_F^2.
    \end{aligned}
\end{eqnarray}
In this subproblem, we only care about whether the real human faces can be warped as the stylized faces with high accuracy, so the term of landmarking error is ignored. 
We solve this problem by Adam~\citep{kingma2014adam}: in each step, we update $\theta$ and $\gamma$ based on a batch of randomly-sampled face pairs. 
\item \textbf{Proximal Face Landmarker Optimization:}
Given $\theta^{(1)}$ and the predicted landmarks (i.e., $\widehat{\bm{Y}}_{i}^{(U)}=f_{\theta^{(1)}}(\bm{X}_i^{(U)})$ for $i=1,..., N_L$), we can treat $\theta^{(1)}$ as the initial variable and optimize it with a proximal regularizer:
\begin{eqnarray}\label{eq:opt_landmarker}
\begin{aligned}
    \theta^{(2)}=&\arg\sideset{}{_{\theta}}\min \sideset{}{_{j=1}^{N_L}}\sum\|f_{\theta}(\bm{X}_j^{(L)})-\bm{Y}_j^{(L)}\|_F^2  \\
    &\quad+\underbrace{\sideset{}{_{i=1}^{N_U}}\sum\|f_{\theta}(\bm{X}_i^{(U)})-\widehat{\bm{Y}}_i^{(U)}\|_F^2}_{\text{Pseudo landmarking error in the target domain}}.
\end{aligned}
\end{eqnarray}
Here, the second term in~\eqref{eq:opt_landmarker} measures the estimation errors of the pseudo landmarks achieved in the previous step.
Essentially, it works as a proximal regularizer, ensuring that the optimized landmarks $f_{\theta^{(2)}}(\bm{X}_i^{(U)})$ is not too far away from the previous estimation $\widehat{\bm{Y}}_i^{(U)}$. 
Similarly, we can solve this problem by Adam~\citep{kingma2014adam} as well. 
\end{itemize}

\begin{figure}[t]
\centering
\includegraphics[width=0.47\textwidth]{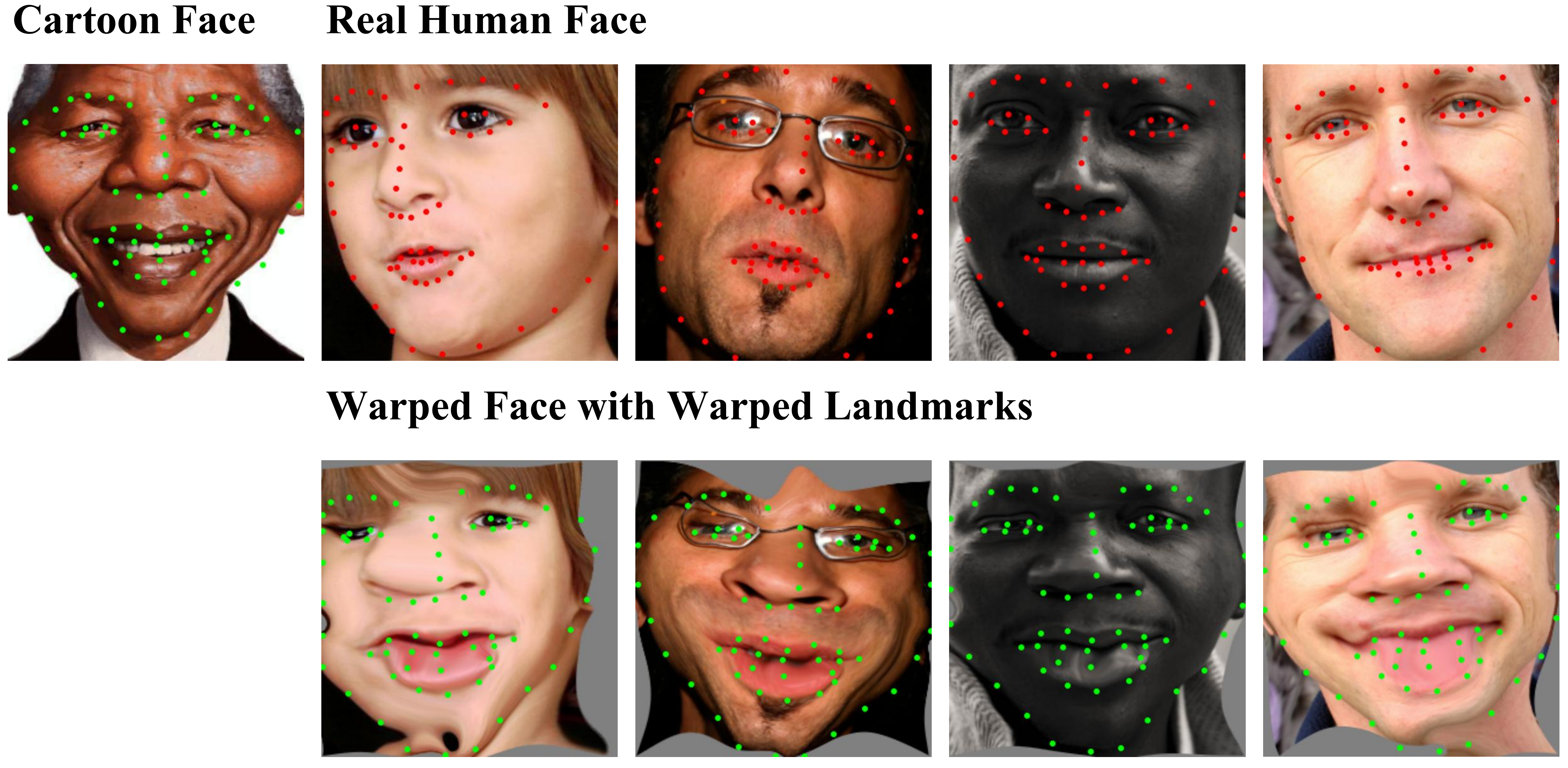}
\caption{Illustrations of conditional face warping results. Taking a cartoon face as the target, our model warps real human faces accordingly. The red dots indicate real human face landmarks, and green dots indicate cartoon and warped face landmarks.}\label{fig: WarpEffect}
\end{figure}

\cref{fig: WarpEffect} shows the warping effect on real human faces achieved by solving~\eqref{eq:opt_warper}.
In \cref{fig: WarpEffect}, the first row shows the real human faces with landmarks and the target stylized face, and the second row shows the warping results, in which the green dots are predicted landmarks. 
These results empirically demonstrate the rationality of our alternating optimization framework.
In particular, we can find that solving~\eqref{eq:opt_warper} leads to reasonable warping results, which are similar to the target stylized face on shape. 
The similarity on face shape indicates that the predicted landmarks can be treated as reliable pseudo labels of the stylized face, which can be used to construct the proximal regularizer that penalizing the pesudo landmarking errors in the target domain. 
Repeating the above two steps till converge, we obtain the target face landmarker that is generalizable for both real and stylized faces.
\cref{alg:learn} shows the learning scheme.

\begin{algorithm}[t]
    \caption{Proposed learning scheme of face landmarker}\label{alg:learn}
	\begin{algorithmic}[1]
	    \REQUIRE 
        Labeled real faces $\mathcal{D}^{(L)}$ and unlabeled stylized faces $\mathcal{D}^{(U)}$.
        The number of iterations (i.e., $M$). 
        Epochs for the subproblems (i.e., $L_1$ and $L_2$).
        \STATE Initialize $\{\gamma^{(0)}\}$ with a pretrained model on $\mathcal{D}^{(L)}$ and $\{\theta^{(0)}\}$ randomly. 
        \FOR{$m=0,...,M-1$}
            \STATE Sample a batch $\{\bm{X}_i^{(U)},\bm{X}_i^{(L)},\bm{Y}_i^{(L)}\}_{i=1}^{N}$.
            \STATE \textbf{Face warper optimization:}
            \STATE Take $\theta^{(2m)}, \gamma^{(m)}$ as the initialization, then solve~\eqref{eq:opt_warper} by Adam with $L_1$ epochs and obtain $\theta^{(2m+1)}$.
            \STATE \textbf{Proximal face landmarker optimization:}
            \STATE Take $\theta^{(2m+1)}$ as the initialization, then solve~\eqref{eq:opt_landmarker} by Adam with $L_2$ epochs and obtain $\theta^{(2m+2)}$.
	    \ENDFOR    
        \RETURN Output a generalizable face landmarker $f_{\theta^{(2M)}}$.
	\end{algorithmic}
\end{algorithm}

\begin{table*}[t]
\caption{Data settings for the three learning paradigms.}\label{tab:dataset}%
\centering
\small{
\tabcolsep=4pt
\begin{tabular}{c|cccccc|ccccc}
\hline\hline
\multirow{2}{*}{Learning} & 
\multicolumn{6}{c|}{\#Training Images and Label Information} & 
\multicolumn{5}{c}{\#Testing Images} \\
\cline{2-12}
\multirow{2}{*}{Paradigm} & 
\multicolumn{2}{c}{\multirow{2}{*}{300W}} & \multicolumn{2}{c}{\multirow{2}{*}{CariFace}} & \multicolumn{2}{c|}{\multirow{2}{*}{ArtiFace}} & \multicolumn{3}{c}{300W} & \multirow{2}{*}{CariFace} & \multirow{2}{*}{ArtiFace} \\ 
& &
& &
& &
& Common & Challenge & Full & & \\
\hline
% Source-only 
% & 3,148 & Labeled 
% & --- & ---  
% & --- & --- 
% & 554 & 135 & 689 & 800 & 160 \\
% \hline
DA (300W$\rightarrow$CariFace) 
& 3,148 & Labeled
& 3,372 & Unlabeled
& --- & --- 
& 554 & 135 & 689 & 800 & --- \\
DA (300W$\rightarrow$ArtiFace) 
& 3,148 & Labeled
& --- & ---
& 128 & Unlabeled 
& 554 & 135 & 689 & --- & 32 \\
\hline
GZSL (Unseen ArtiFace)
& 3,148 & Labeled
& 3,372 & Unlabeled
& --- & --- 
& 554 & 135 & 689 & 800 & 160 \\
GZSL (Unseen CariFace)
& 3,148 & Labeled
& --- & ---
& 128 & Unlabeled 
& 554 & 135 & 689 & 800 & 32 \\
\hline
Oracle 
& 3,148 & Labeled
& 3,372 & Labeled
& 128 & Labeled
& 554 & 135 & 689 & 800 & 32\\
\hline\hline
\end{tabular}
}
\end{table*}

\section{Experiment}
We apply our learning method to learn a face landmarker and test it on landmarking faces with various styles. 
Extensive experiments, including comparisons with baselines and analytic ablation studies, demonstrate the effectiveness of our learning method and the generalizability of the corresponding face landmarker. 
All the experiments are conducted on a single NVIDIA 3090 GPU. 
Representative experimental results are shown below. 
\textbf{More experimental results and implementation details are given in the supplementary material.}

\subsection{Dataset}
\begin{figure}[t]
\centering
\includegraphics[width=1\linewidth]{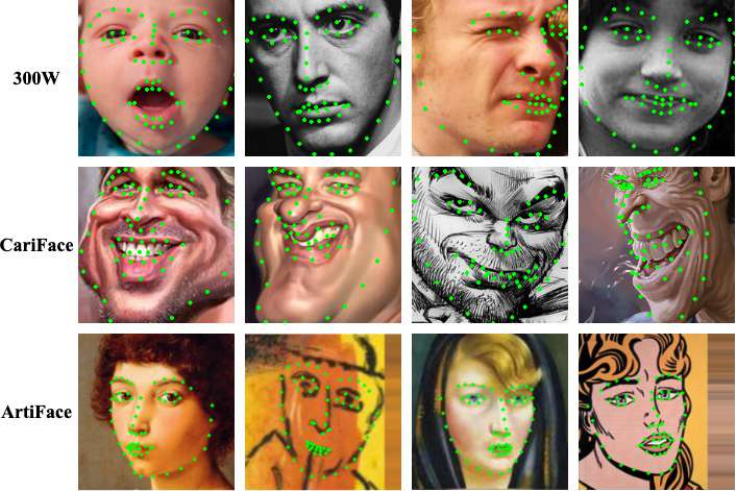}
\caption{
Illustrations of typical samples in the 300W, CariFace, and ArtiFace datasets, each of which is annotated with landmarks.
}
\label{fig:dataset}
\end{figure}

In this study, we conduct experiments based on the following three commonly-used face datasets.
\begin{itemize}
    \item \textbf{300W Dataset.} 300W~\citep{Sagonas_Tzimiropoulos_Zafeiriou_Pantic_2014} is comprised of five well-known real human face datasets including LFPW~\citep{Belhumeur_Jacobs_Kriegman_Kumar_2011}, AFW~\citep{zhu2012face}, HELEN~\citep{Le_Brandt_Lin_Bourdev_Huang_2012}, XM2VTS~\citep{messer1999xm2vtsdb}, and IBUG~\citep{Sagonas_Tzimiropoulos_Zafeiriou_Pantic_2014}.
    \item \textbf{CariFace Dataset.} CariFace~\citep{Cai_Guo_Peng_Zhang_2020} is created by searching and selecting thousands of various caricatures from different celebrities on the Internet.
    \item \textbf{ArtiFace Dataset.} ArtiFace~\citep{yaniv2019face} contains 160 artistic portraits of 16 artists, which covers diverse artwork styles ranging from Renaissance to Comics.
\end{itemize}
Each face in the datasets is annotated with 68 landmarks. 
Typical faces in the datasets and their landmarks are shown in~\cref{fig:dataset}. 
We can find that the faces in the three datasets have distinguished styles, which correspond to three different domains. 
In particular, compared to 300W~\citep{Sagonas_Tzimiropoulos_Zafeiriou_Pantic_2014}, CariFace~\citep{Cai_Guo_Peng_Zhang_2020} exhibits abstract and exaggerated patterns, leading to large representation variations. 
ArtiFace~\citep{yaniv2019face} not only has larger variations across different artistic categories but also differs greatly in terms of the aspect of facial scales, orientations, locations, and so on.

% We conduct experiments on widely-used 300W dataset \citep{Sagonas_Tzimiropoulos_Zafeiriou_Pantic_2014} and CariFace dataset \citep{Cai_Guo_Peng_Zhang_2020}. 

% Additionally, 2,000 caricatures are generated using two Generative Adversarial Networks (GANs).
% Among the initial 6,000 caricatures, 68 landmarks are manually labeled.
% In the case of the generated caricatures, their landmarks are automatically generated in correspondence with the artistic features.
% % Furthermore, we access the generalization ability of the proposed method on unseen categories by testing it on the 
% % and Anime dataset \citep{branwen2019danbooru2019} which is a large-scale anime head illustration dataset. 
% % In our study, we select a subset of 1,000 images from the dataset and utilize the anime landmark predictor \citep{anime-face-detector} to generate 28 landmarks for each image. 
% % Note the anime dataset contains only 28 keypoints, as opposed to the 68 keypoints in the common settings, we established a corresponding relationship between them, as illustrated in Fig.~\ref{fig:relationship between 28 to 68}. 

\subsection{Learning Paradigms and Baselines}
Given the above datasets, we consider the following three learning paradigms:
\begin{itemize}
    % \item \textbf{Source-only.} We learn a face landmarker merely based on the 300W real human faces and evaluate its performance on the other two datasets.
    \item \textbf{Domain Adaptation (DA).} Given labeled 300W faces and unlabeled stylized faces from CariFace or ArtiFace, we learn a face landmarker based on various domain adaptation methods. 
    \item \textbf{Generalized Zero-shot Learning (GZSL).}
    In the challenging GZSL setting, we learn a face landmarker based on the above DA-based methods and test it in an unseen face domain (e.g., learning the landmarker on labeled 300W and unlabeled CariFace and testing on ArtiFace). 
    \item \textbf{Oracle.} In this setting, the labeled faces of all three datasets are accessible, and we can learn the face landmarker by classic supervised learning. 
\end{itemize}
For a fair comparison, in each learning paradigm, we set the architecture of the face landmarker based on the SLPT in~\citep{xia2022sparse}. 
Ideally, we would like to learn landmarkers in the DA and GZSL settings, making its performance comparable to the oracle. 
In the oracle setting, we can learn the face landmarker directly via classic supervised learning (SL), i.e., $\min_{\theta}\sum_{(\bm{X},\bm{Y})\sim \mathcal{D}}\|f_{\theta}(\bm{X})-\bm{Y}\|_F^2$. 
In the DA and GZSL settings, besides minimizing the landmark estimation errors, we can apply various image style transfer and domain adaptation methods, e.g., RevGrad~\citep{ganin2015unsupervised}, CycleGAN~\citep{Zhu_2017_ICCV}, BDL~\citep{li2019bidirectional}, AdaptSegNet~\citep{tsai2018learning} and FDA~\citep{yang2020fda}, to impose domain adaptation regularization during training. 
These methods work as the baselines of our method.

\begin{table*}[t]
\caption{Comparisons for various methods on their NMEs. In DA and GZSL settings, the best results are bold.}\label{tab:cmp}
\centering
\small{
\begin{tabular}{c|l|ccccc|c}
\hline\hline
\multirow{2}{*}{Learning Paradigm} & \multirow{2}{*}{Learning Method} &
\multicolumn{3}{c}{300W} & \multirow{2}{*}{CariFace} & \multirow{2}{*}{ArtiFace} & Average \\ 
& & Common & Challenge & Full & & & NME\\
\hline
% Source-only & SL &2.75 &4.90 &3.17 & 11.05 &4.56 &7.13 \\
% \hline
 & SL+RevGrad~\citep{ganin2015unsupervised}  
&2.84 &5.58 & 3.38&12.19 &5.16 & 7.83 \\
 \multirow{2}{*}{DA (300W$\rightarrow$CariFace)}& 
SL+CycleGAN~\citep{Zhu_2017_ICCV} 
&\textbf{2.74} &5.43 &3.27 &12.11 &4.70 &7.70 \\
\multirow{2}{*}{and}& SL+BDL~\citep{li2019bidirectional}  
& 3.28 & 6.17 & 3.84 & 13.63 & 5.65 & 8.77  \\
\multirow{2}{*}{GZSL (Unseen ArtiFace)}& SL+ASN~\citep{tsai2018learning}
&2.92 &5.26 &3.38 &12.21 & 4.75 & 7.80 \\
& SL+FDA~\citep{yang2020fda}
&2.89 &5.18 &3.34 &12.66 & 4.60 & 8.07 \\
& Ours 
&2.79 &\textbf{4.91} &\textbf{3.20} &\textbf{7.70} & \textbf{3.95} &\textbf{5.46}  \\
\hline
& SL+RevGrad~\citep{ganin2015unsupervised}  
&2.99 &5.81 &3.55 &12.46 &4.74 & 8.26  \\
\multirow{2}{*}{DA (300W$\rightarrow$ArtiFace)}& SL+CycleGAN~\citep{Zhu_2017_ICCV}  
&3.00 &5.65 &3.52 &12.64 &5.34 & 8.36 \\
\multirow{2}{*}{and}& SL+BDL~\cite{li2019bidirectional}  
& 2.99 & 5.32 &3.44 & 13.40 & 5.90 & 8.73  \\
\multirow{2}{*}{GZSL (Unseen CariFace)}& SL+ASN~\citep{tsai2018learning}
&\textbf{2.89} &5.81 &3.46 & 16.58 &5.65 &10.31  \\
& SL+FDA~\citep{yang2020fda}
&3.05 &6.21 &3.68 &12.33 &5.87 &  8.27\\
& Ours 
&2.90 & \textbf{5.14} & \textbf{3.34} & \textbf{10.93} & \textbf{3.93} & \textbf{7.34} \\
\hline
Oracle & SL
&2.68 &4.86 &3.10 & 5.48 & 3.31 & 4.36  \\
\hline\hline
\end{tabular}
}
\end{table*}

Given the landmarkers learned by various methods, we evaluate them with the standard metric, Normalized Mean Error (NME).
In \cref{tab:dataset}, we show the training and testing data settings in the above three learning paradigms. 
Following existing work~\citep{dong2018supervision, wang2020deep, xia2022sparse}, we further split the 689 testing faces in 300W into 554 faces in common scenarios and 135 faces in challenging scenarios. 
The NMEs for the common, challenge, and full scenarios are recorded.

\begin{figure}[t]
    \centering
    \begin{subfigure}[b]{1\linewidth}
        \centering
        \includegraphics[height=1.6cm]{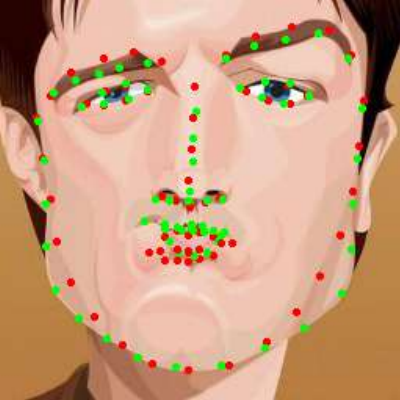}
        \includegraphics[height=1.6cm]{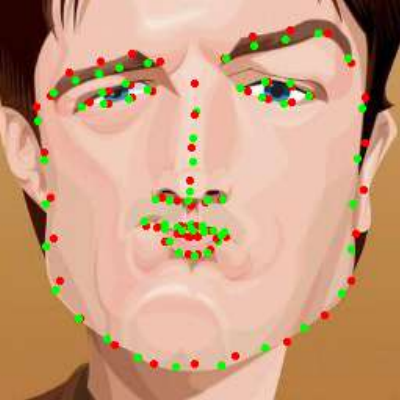}
        \includegraphics[height=1.6cm]{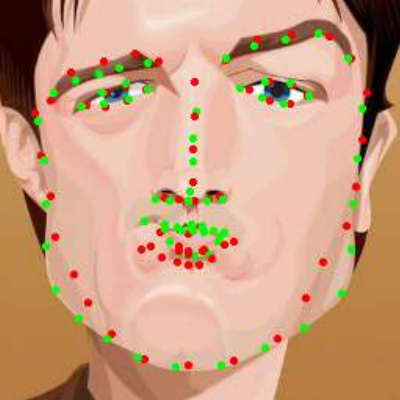}
        \includegraphics[height=1.6cm]{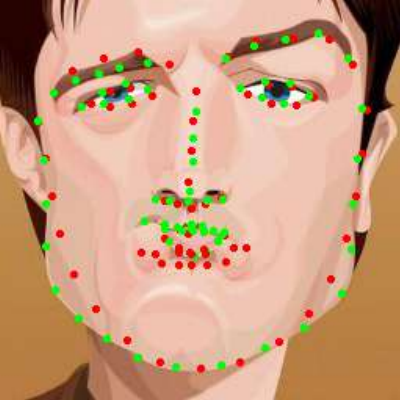}
        \includegraphics[height=1.6cm]{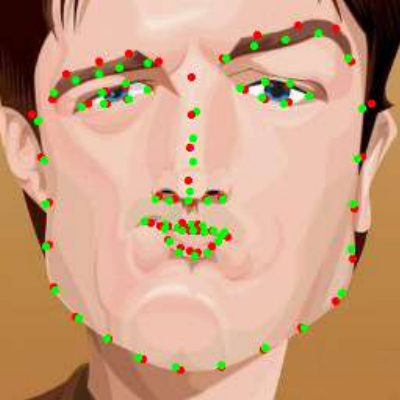}\\
        \includegraphics[width=1.6cm,height=0.7cm]{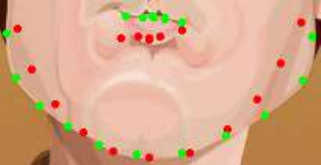}
        \includegraphics[width=1.6cm,height=0.7cm]{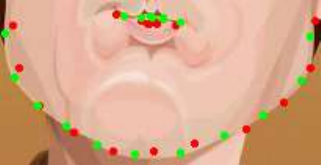}
        \includegraphics[width=1.6cm,height=0.7cm]{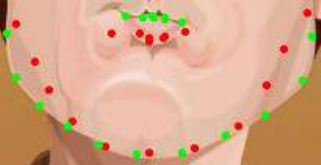}
        \includegraphics[width=1.6cm,height=0.7cm]{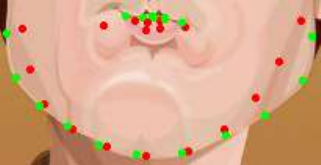}
        \includegraphics[width=1.6cm,height=0.7cm]{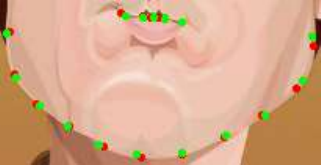}\\
        \small{Source-only\hspace{0.3cm} RevGrad\hspace{0.4cm}CycleGAN\hspace{0.6cm} BDL\hspace{0.9cm} Ours\hspace{0.5cm}}
        \subcaption{DA (300W$\rightarrow$CariFace)}
    \end{subfigure}
    \begin{subfigure}[b]{1\linewidth}
        \centering
         \includegraphics[height=1.6cm]{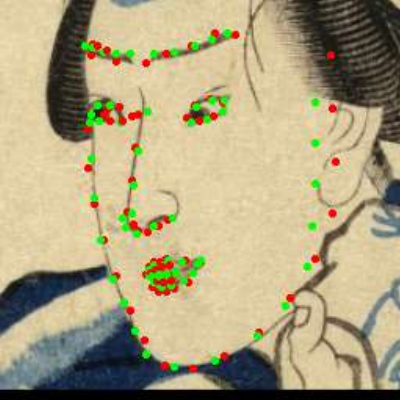}
        \includegraphics[height=1.6cm]{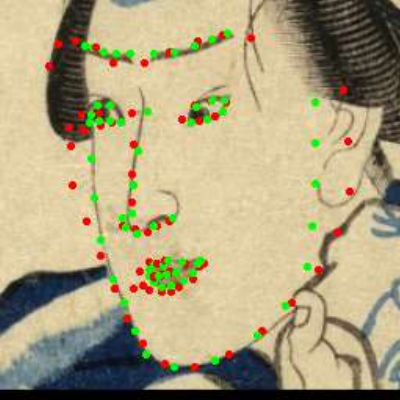}
        \includegraphics[height=1.6cm]{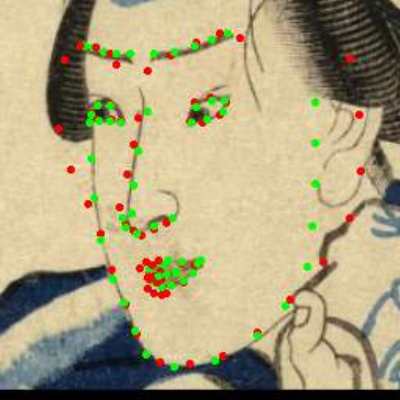}
        \includegraphics[height=1.6cm]{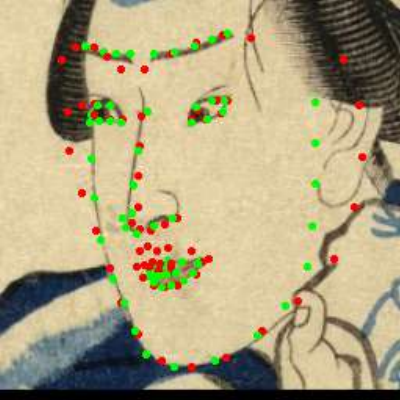}
        \includegraphics[height=1.6cm]{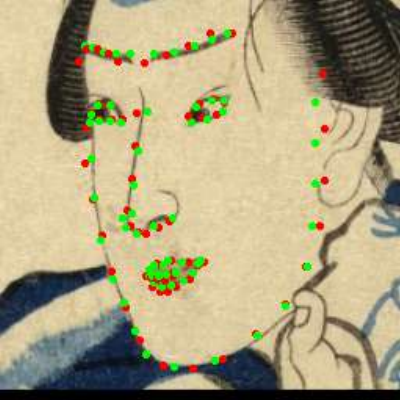}\\
        \includegraphics[width=1.6cm]{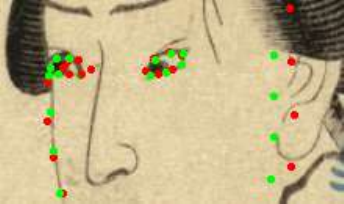}
        \includegraphics[width=1.6cm]{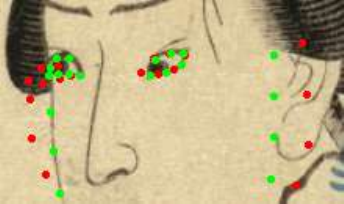}
        \includegraphics[width=1.6cm]{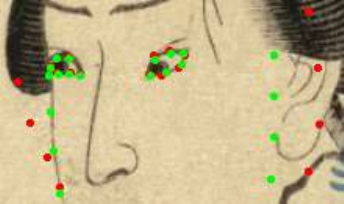}
        \includegraphics[width=1.6cm]{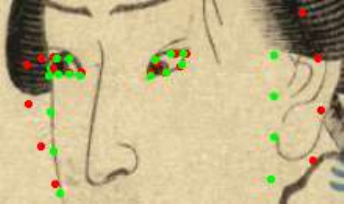}
        \includegraphics[width=1.6cm]{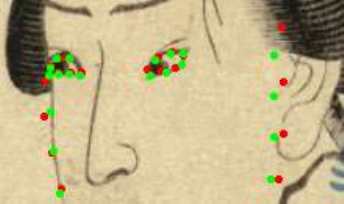}\\
        \small{Source-only\hspace{0.3cm} RevGrad\hspace{0.4cm}CycleGAN\hspace{0.6cm} BDL\hspace{0.9cm} Ours\hspace{0.5cm}}
        \subcaption{DA (300W$\rightarrow$ArtiFace)}
    \end{subfigure}
    \caption{Visual comparisons for various methods in the two DA settings. 
    We only highlight points on the inner lips in the enlarged region of the mouth in (a), as well as the eyes and the sides of the cheeks, excluding points on the eyebrows in (b).}
    \label{fig:cmp_da}
\end{figure}

\begin{figure}[t]
    \centering
    \begin{subfigure}[b]{1\linewidth}
        \centering
        \includegraphics[height=1.6cm]{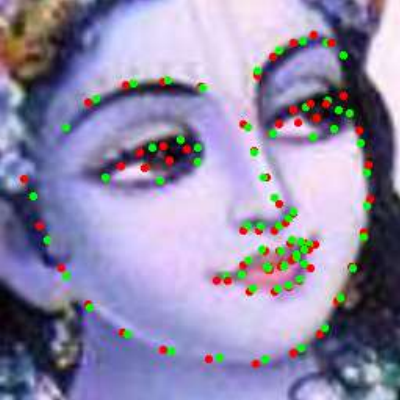}
        \includegraphics[height=1.6cm]{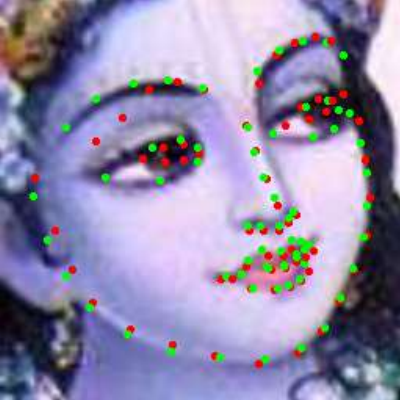}
        \includegraphics[height=1.6cm]{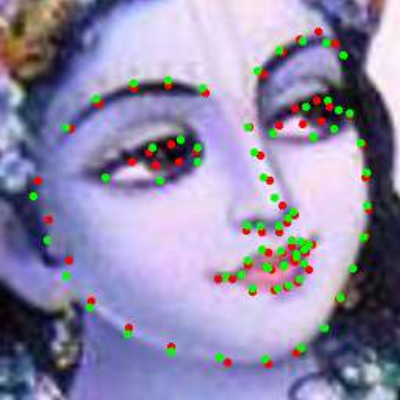}
        \includegraphics[height=1.6cm]{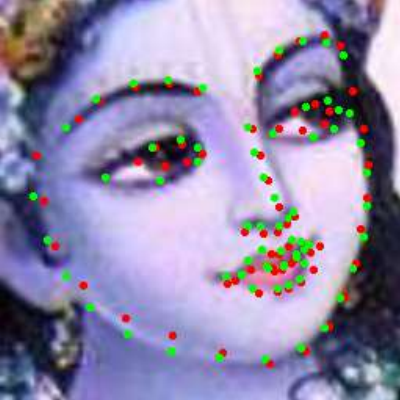}
        \includegraphics[height=1.6cm]{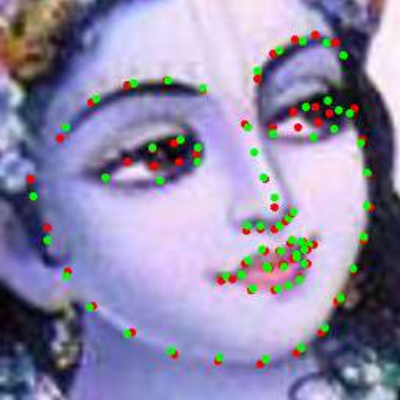}\\
       \includegraphics[width=1.6cm]{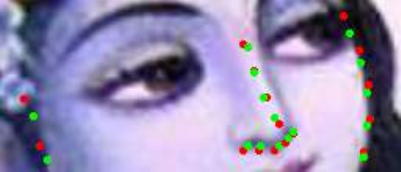}
        \includegraphics[width=1.6cm]{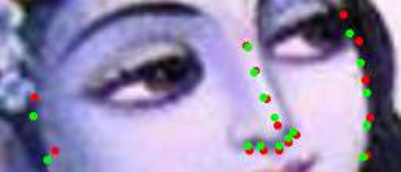}
        \includegraphics[width=1.6cm]{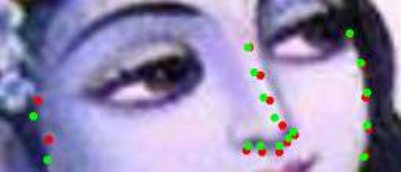}
        \includegraphics[width=1.6cm]{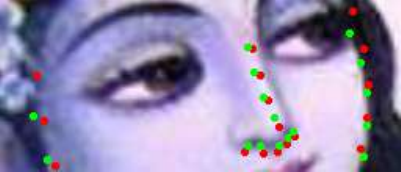}
        \includegraphics[width=1.6cm]{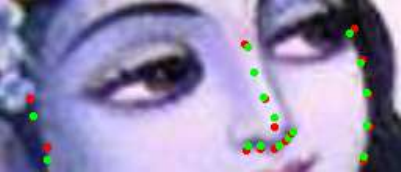}\\
        \small{Source-only\hspace{0.3cm} RevGrad\hspace{0.4cm}CycleGAN\hspace{0.6cm} BDL\hspace{0.9cm} Ours\hspace{0.5cm}}
        \subcaption{Case 1}
    \end{subfigure}
    \begin{subfigure}[b]{1\linewidth}
        \centering
       \includegraphics[height=1.6cm]{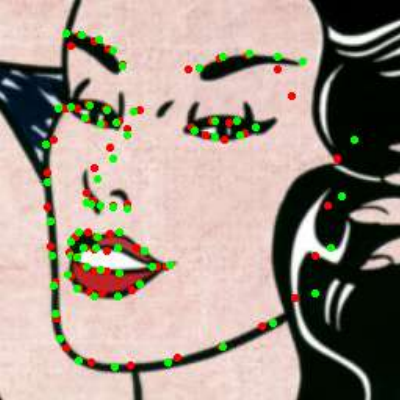}
        \includegraphics[height=1.6cm]{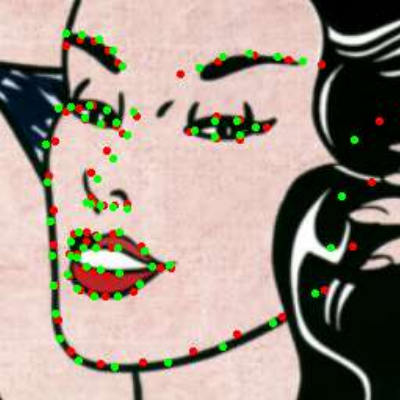}
        \includegraphics[height=1.6cm]{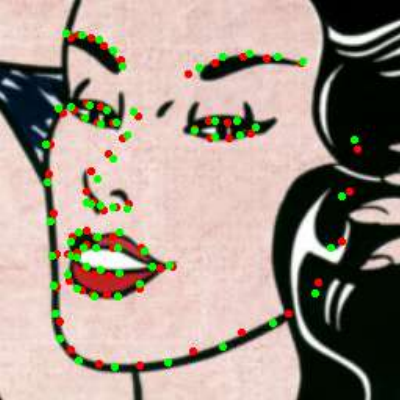}
        \includegraphics[height=1.6cm]{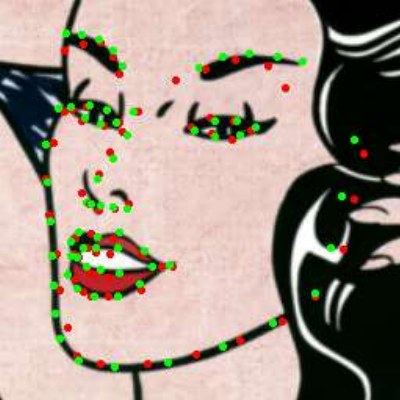}
        \includegraphics[height=1.6cm]{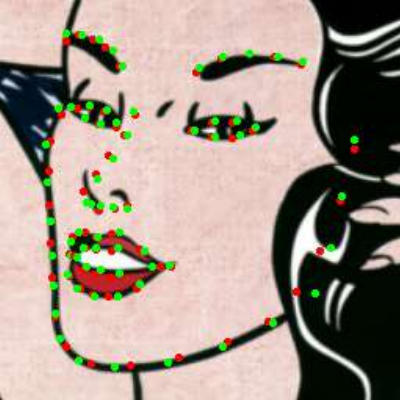}\\
       \includegraphics[width=1.6cm]{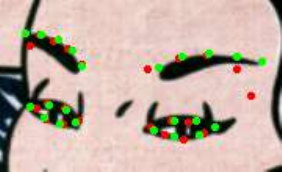}
        \includegraphics[width=1.6cm]{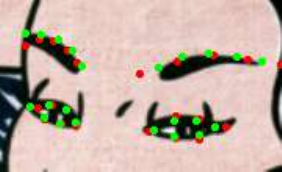}
        \includegraphics[width=1.6cm]{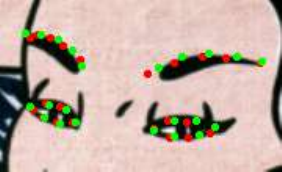}
        \includegraphics[width=1.6cm]{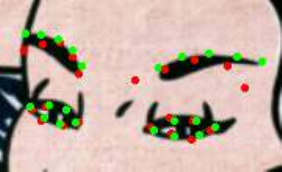}
        \includegraphics[width=1.6cm]{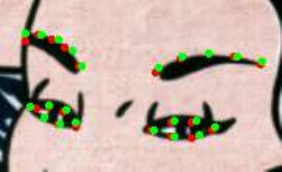}\\
        \small{Source-only\hspace{0.3cm} RevGrad\hspace{0.4cm}CycleGAN\hspace{0.6cm} BDL\hspace{0.9cm} Ours\hspace{0.5cm}}
        \subcaption{Case 2}
    \end{subfigure}
    \caption{Visual comparisons for various methods in the GZSL (Unseen ArtiFace) setting.
     The zoomed-in area in (a) highlights the noses and facial contours, while that in (b) concentrates on the upper part of faces.}
    \label{fig:cmp_gzsl}
\end{figure}

\begin{figure*}[t]
    \centering
    \begin{subfigure}[b]{0.19\textwidth}
        \centering
        \includegraphics[height=3.5cm]{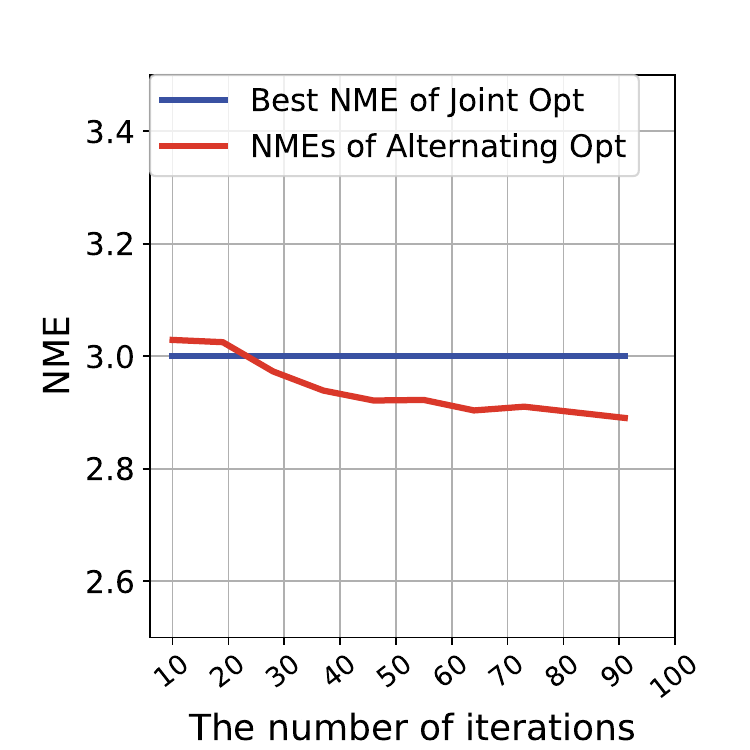}
        \subcaption{300W (Common)}
    \end{subfigure}
    \begin{subfigure}[b]{0.19\textwidth}
        \centering
        \includegraphics[height=3.5cm]{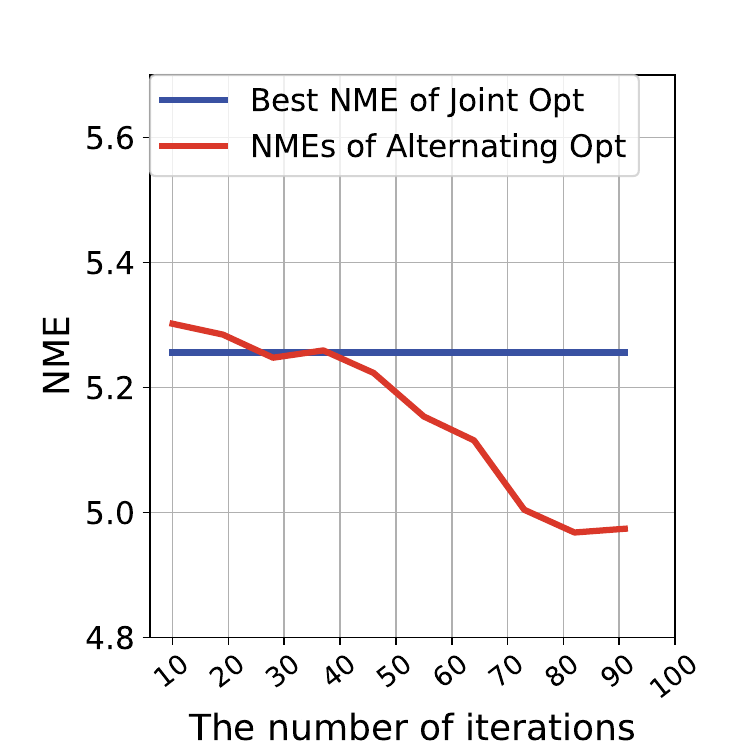}
        \subcaption{300W (Challenge)}
    \end{subfigure}
    \begin{subfigure}[b]{0.19\textwidth}
        \centering
        \includegraphics[height=3.5cm]{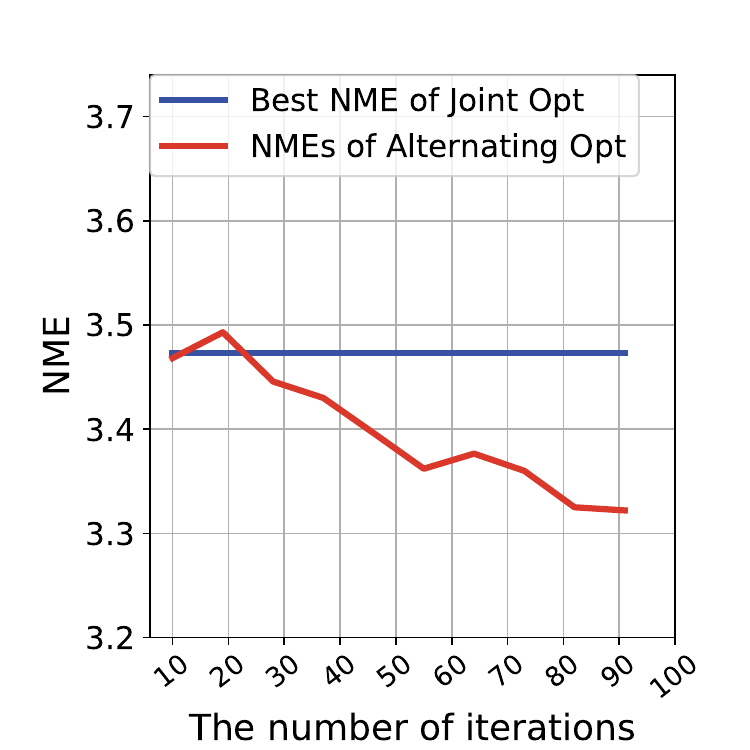}
        \subcaption{300W (Full)}
    \end{subfigure}
    \begin{subfigure}[b]{0.19\textwidth}
        \centering
        \includegraphics[height=3.5cm]{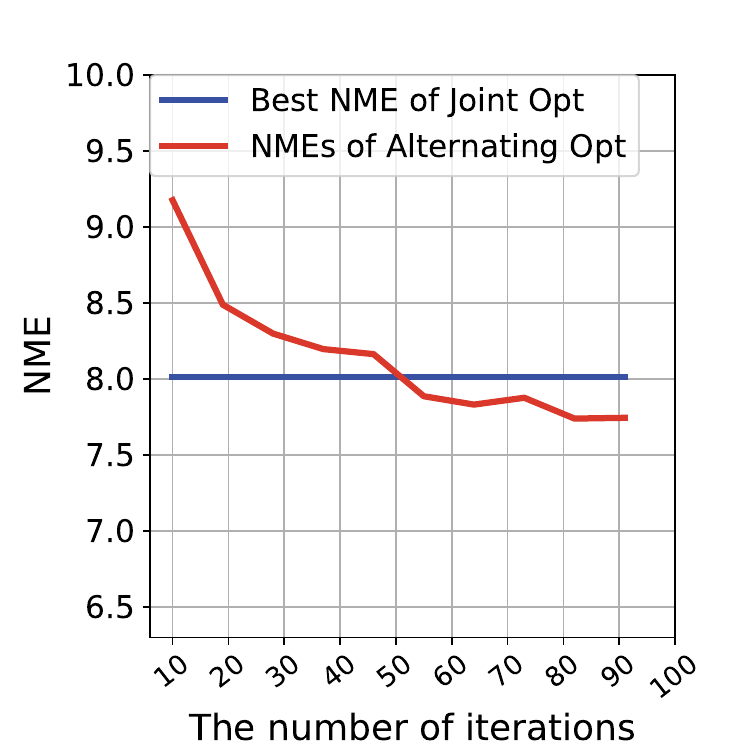}
        \subcaption{CariFace (Target)}
    \end{subfigure}
    \begin{subfigure}[b]{0.19\textwidth}
        \centering
        \includegraphics[height=3.5cm]{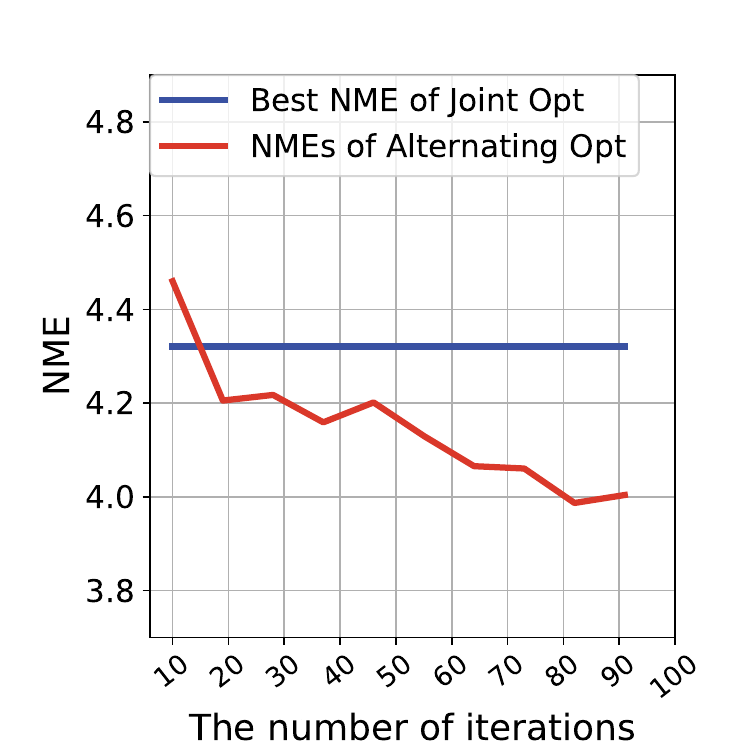}
        \subcaption{ArtiFace (Unseen)}
    \end{subfigure}
    \caption{Comparisons for different optimization strategies on NME performance in the GZSL (Unseen ArtiFace) setting.}
    \label{fig:opt}
\end{figure*}

\begin{table*}[t]
\caption{The impacts of different model architectures on the NME performance of our method. In each setting, the best results are bold.}\label{tab:model}
\centering
\small{
\begin{tabular}{c|l|ccccc|c}
\hline\hline
\multirow{2}{*}{Learning Paradigm} & \multirow{2}{*}{Learning Method} &
\multicolumn{3}{c}{300W} & \multirow{2}{*}{CariFace} & \multirow{2}{*}{ArtiFace} & Average \\ 
& & Common & Challenge & Full & & & NME\\
\hline
DA (300W$\rightarrow$CariFace) & SBR~\citep{dong2018supervision}
&3.61 & 6.36 &4.15 &8.00& 5.09&6.11 \\
and & HRNet~\citep{wang2020deep}  
&2.93  &5.29  & 3.40 &7.73 &4.33 & 5.60 \\
GZSL (Unseen ArtiFace) & SLPT~\citep{xia2022sparse} (Default)  
&\textbf{2.79} &\textbf{4.91} &\textbf{3.20} &\textbf{7.70} & \textbf{3.95} &\textbf{5.46}  \\
\hline
DA (300W$\rightarrow$ArtiFace) &SBR~\citep{dong2018supervision}  &3.70 &6.82 &4.32 &11.35 &5.32 &8.04  \\
and & HRNet~\citep{wang2020deep}  
&3.04 &5.44 &3.51 &\textbf{9.85} &4.30 & \textbf{6.86} \\
GZSL (Unseen CariFace)& SLPT~\citep{xia2022sparse} (Default)  
&\textbf{2.90} & \textbf{5.14} & \textbf{3.34 }& 10.93 & \textbf{3.93} & 7.34  \\
\hline\hline
\end{tabular}
}
\end{table*}

\subsection{Numerical and Visual Comparisons}
In \cref{tab:cmp}, we show the performance of various learning methods in different settings, demonstrating the effectiveness and superiority of our method.
% In particular, when training the SLPT face landmarker supervisedly in the source-only setting, the model performs well on 300W but suffers from poor generalizability on other datasets.
In particular, existing DA methods often fail to improve the generalization power of model a lot in face landmarking tasks --- their performance in target and unseen domains is inferior to that in the oracle setting, with a significant gap on NME. 
A potential reason for this phenomenon is that these methods focus on the adaptation of image domain and the landmark-related loss is not dominant in their learning processes.
As a result, instead of learning the face landmarker, they make more efforts to optimize the parameters of other modules (e.g., the neural network-based face stylization modules and discriminators) during training. 

Different from the baselines, the performance of our model in the target domain (e.g., CariFace and ArtiFace) is improved consistently, while the performance in the source domain does not degrade a lot. 
In both DA and GZSL settings, our method outperforms the baselines in most situations.
Especially in the GZSL settings, the landmarkers obtained by our method show encouraging generalization power, which achieves lower NME than the baselines in the unseen domain (i.e., ArtiFace). 
Overall, the NME of our method in the source domain (i.e., 300W) is comparable to that achieved in the oracle setting. 
For the target even unseen domains, our method reduces the gap to the oracle. 

Besides the numerical comparisons, we provide some visualization results obtained by different methods in~\cref{fig:cmp_da,fig:cmp_gzsl}.
For a complete comparison, the results of the SLPT trained only on 300W are shown in the figures as well, which corresponds to learning a face landmarker only based on the data in the source domain.
We can find that by applying our learning method, the face landmarker obtains better landmarks, which aligns with the ground truth with smaller errors, particularly for the landmarks of face contour, nose, mouth, and eyebrow. 

To further verify the generalization of our method, we select three more state-of-the-art landmarkers, including OpenPose (OP)~\citep{cao2017realtime}, SPIGA~\citep{prados2022shape} and STAR~\citep{zhou2023star} for numerical comparisons, and incorporate STAR as the backbone model of our method.
\cref{sota_method} demonstrates that our learning method is applicable for various backbone models (e.g., STAR and SLPT), and integrating existing landmarkers with our learning method is an effective and competitive approach to enhance their generalizability.
\begin{table}[t]
\caption{Comparison with SOTA landmark detectors on NME.}\label{sota_method}
\centering
\tabcolsep=3pt
\small{
\begin{tabular}{c|cccc|cc}
% \toprule%
\hline\hline
% {Model} & Openpose ~\citep{cao2017realtime} & SPFL ~\citep{prados2022shape}& STAR ~\citep{zhou2023star} & Ours  \\
Learning Method & \multicolumn{4}{c|}{Train on 300W} & \multicolumn{2}{c}{Our method}\\
\hline
{Backbone} & OP & SPIGA & STAR & SLPT & STAR  & SLPT  \\
% \midrule
\hline
DA on {Cariface} & 10.46 & 11.23 & 10.97 & 11.05 & 7.62 & 7.70 \\
GZSL on {Artiface} & 5.55& 5.19 & 5.30 & 4.56 & 4.86 & 3.93\\
% \bottomrule
\hline\hline
\end{tabular}
}
\end{table}

\subsection{Analytic Experiments}
% To further demonstrate the rationality of our model and learning algorithm, we conduct the following ablation studies and quantitatively analyze the impacts of different model/algorithmic configurations on our learning results.

\subsubsection{Joint v.s. Alternating Optimization}\label{sec:effect on training mode}
As aforementioned, learning the face landmarker and the warping field model jointly makes it much easier to fall into undesired local optimum or unstable saddle points.
To verify this claim, we apply the joint optimization strategy, i.e., solving~\eqref{eq:opt} by optimizing $\theta$ and $\gamma$ jointly in each gradient descent step and compare its performance with ours. 
In \cref{fig:opt}, we show the best NME achieved by this joint optimization strategy and the NME achieved by our alternating optimization method in each iteration. 
We can find that the joint optimization strategy tends to overfit the source domain (300W), leading to power generlizability in the target and unseen domains.
On the contrary, with the increase of iterations, the NMEs of our method on the target and unseen domains decrease consistently and become lower than those of the joint optimization after several iterations.

\subsubsection{Impacts of Model Architectures}\label{sec: backbone}
Besides the optimization strategy, the model architecture also has an impact on the model performance. 
By default, we implement the face landmarker as the SLPT model. 
In this experiment, we further explore the performance of other model architectures, including SBR~\citep{dong2018supervision} and HRNet~\citep{wang2020deep}.
\cref{tab:model} shows the performance of different model architectures in DA and GZSL settings. 
We can find that the SLPT-based face landmarker works better than the two competitors in this experiment. 
A potential reason for this phenomenon is that both SBR and HRNet are heatmap-based face landmarkers. 
Given an input facial image, they output a heatmap indicating the distribution of landmarks rather than a set of deterministic landmark coordinates.
We have to first detect the landmarks from the heatmap and then pass them through the warping field model.
Accordingly, in the face warper optimization step (i.e., solving~\eqref{eq:opt_warper}), the landmarker and the warping field model cannot be trained in an end-to-end way because the backpropagation of the gradient becomes inapplicable. 
As a result, we have to update $\theta$ and $\gamma$ alternatively, leading to suboptimal performance.

\subsubsection{Effects on Loss Term}

In the context of our model's loss function, the image gradient field serves as its input. 
It has been observed through experiments that utilizing distinct gradient operators leads to the generation of varying image gradient fields, which in turn influences the model's performance. 
Consequently, in this section, we assess and compare the training outcomes derived from image gradient fields obtained through the application of diverse gradient operators.

Moreover, it is crucial to recognize that the choice of the input image, specifically whether it is a grayscale image or not, can also exert a certain influence on the results. 
As a result, we have incorporated a comparison of the outcomes when the input is a grayscale image.
It is important to emphasize that all experiment settings are maintained consistently.
The results are shown in \cref{ablation_loss}.
Utilizing grayscale images yields superior effects compared to not using grayscale images. 
This could be attributed to the fact that converting an image to grayscale eliminates the trivial effects of color and complex textures. 
The most optimal outcome is achieved by employing the Sobel operator. 
This may be due to the Sobel operator's emphasis on edge information within the image, enabling it to capture more geometric details. 
Additionally, the Sobel operator incorporates a mild smoothing effect during gradient computation, which mitigates the impact of noise. 

\begin{table}[t]
\caption{Quantitative comparison on CariFace dataset.}\label{ablation_loss}
\centering 
\small{
\begin{tabular}{lccccc}
% \toprule%
\hline\hline
\multirow{1}{*}{\text{Operator}} &\multirow{1}{*}{Gray} &\multirow{1}{*}{Spatial} &\multirow{1}{*}{Laplacian} &\multirow{1}{*}{Canny} &\multirow{1}{*}{Sobel}  \\
\midrule
\multirow{2}{*}{\text{NME}} & \checkmark & 8.282 & 7.870 & 8.005 & \textbf{7.695}  \\
 & $\times$ & 7.958 & 7.921 & 8.011 & 7.863 \\
% \bottomrule
\hline\hline
\end{tabular}
}
\end{table}

Furthermore, we consider $i)$ replacing the gradient MSE loss in~\eqref{eq:opt} with the pixel MSE or perceptual loss and $ii)$ removing the gradient MSE loss. 
\cref{tab: different_loss} shows the rationality of the gradient MSE for the reason that $i)$ landmarks are distributed on edges; and $ii)$ the gradient field filters out unnecessary color information, simplifying the task. 
These findings form the basis for our decision to adopt this particular method in our experimental process.

\begin{table}[t]
% \vspace{-10pt}
\caption{Comparisons for different losses on NME.}\label{tab: different_loss}
% \vspace{-10pt}
\centering
\tabcolsep=2pt
\small{
\begin{tabular}{c|c|c|c|c|c}
\hline\hline%
\multicolumn{2}{c|}{{Type of Loss}} & MSE & Perceptual & w.o. Grad MSE & Grad MSE\\
\hline
\multirow{3}{*}{300W} & Common & 3.08 & 2.89  & 2.96 &\textbf{2.79}\\
& Challenge & 5.43 & 5.13 & 5.00 & \textbf{4.91}\\
& Full & 3.54 & 3.33 & 3.36&  \textbf{3.20}\\
\hline
\multicolumn{2}{c|}{DA on Caricature} & 8.26 & 8.23  & 9.48& \textbf{7.70}\\
\hline
\multicolumn{2}{c|}{GZSL on ArtiFace} & 4.12 & 4.07  & 4.06& \textbf{3.95}\\
\hline\hline
\end{tabular}
}
\end{table}

\subsubsection{Rationality of Proposed Warping Field Model}
\begin{figure}[t]
    \centering
    \begin{subfigure}[b]{0.32\linewidth}
        \centering
        \includegraphics[height=2.4cm]{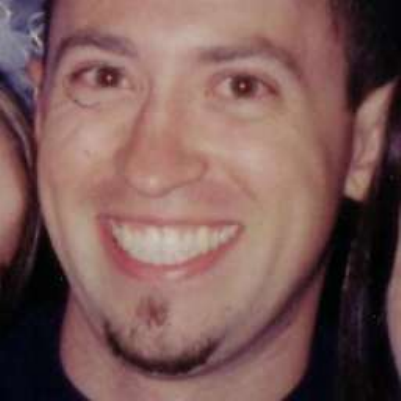}
        \subcaption{Source}
    \end{subfigure}
    \begin{subfigure}[b]{0.32\linewidth}
        \centering
        \includegraphics[height=2.4cm]{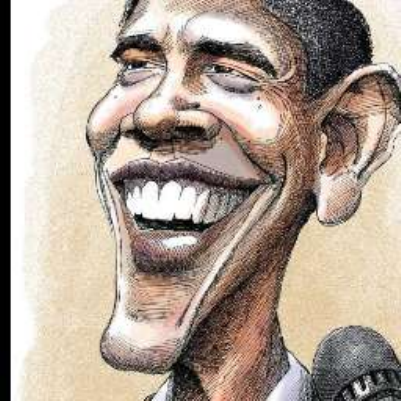}
        \subcaption{Target}
    \end{subfigure}
    \begin{subfigure}[b]{0.32\linewidth}
        \centering
        \includegraphics[height=2.4cm]{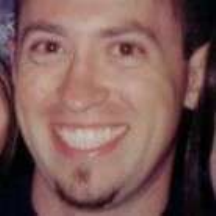}
        \subcaption{AutoToon}
    \end{subfigure}\\
    \begin{subfigure}[b]{0.32\linewidth}
        \centering
        \includegraphics[height=2.4cm]{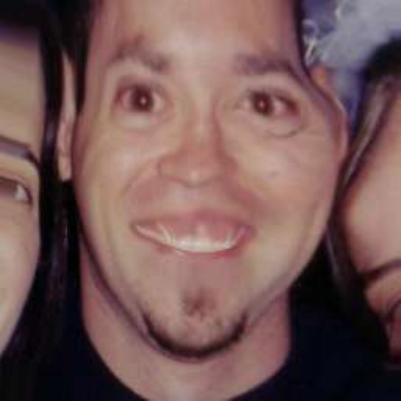}
        \subcaption{CariGANs}
    \end{subfigure}
    \begin{subfigure}[b]{0.32\linewidth}
        \centering
        \includegraphics[height=2.4cm]{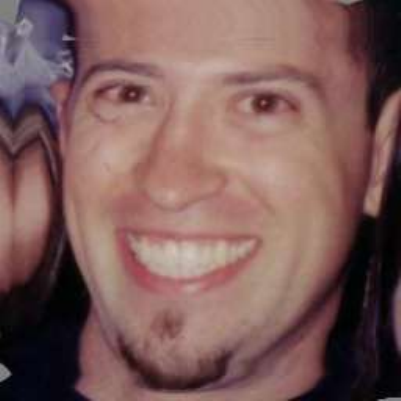}
        \subcaption{RAFT}
    \end{subfigure}
    \begin{subfigure}[b]{0.32\linewidth}
        \centering
        \includegraphics[height=2.4cm]{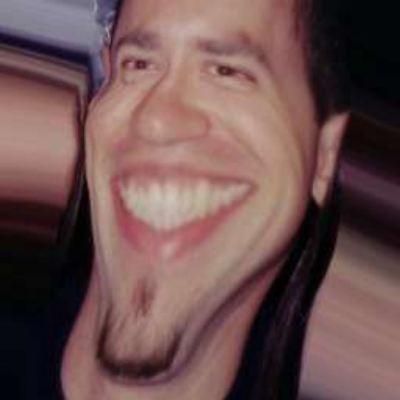}
        \subcaption{Ours}
    \end{subfigure}
    \caption{Comparisons for different face warpers.}
    \label{fig:warp}
\end{figure}

Besides the model architecture of the face landmarker, the warping field model impacts our face landmarker as well --- when the predicted warping field is inaccurate, we cannot obtain reliable pseudo landmarks for stylized faces. 
Therefore, we investigate different face warping models and demonstrate the rationality of the warping field model implemented in our work.
In particular, applying different warping field models in our training process, we visualize their warping results in \cref{fig:warp}. 
We can find that although the face landmarker with the polyharmonic interpolation model leads to a very simple face warper, it outperforms many existing neural network-based image warping methods, such as AutoToon~\cite{gong2020autotoon} and CariGANs~\cite{cao2018carigans} for facial manipulation and the optical flow methods like RAFT~\citep{teed2020raft}.
These methods either require one-to-one correspondence information between source and target images or assume the deformation between the two images to be slight, making them unsuitable for our problem, especially for the stylized faces with significant nonrigid deformations. 

\section{Conclusion}
In this paper, we propose a simple but effective method for learning a generalizable face landmarker applicable to facial images with different styles.
Given labeled real human faces and unlabeled stylized faces, our method learns the face landmarker under the guidance of conditional face warping, demonstrating the usefulness of the warping information. 
An alternating optimization framework is proposed to learn the face landmarker together with the warping field model.
Experiments demonstrate the effectiveness of our method. Especially in the generalized zero-shot learning scenarios, our method achieves encouraging landmarking accuracy in unseen face domains.

\section*{Acknowledgements}
This work was supported in part by the National Natural Science Foundation of China (62102031), the foundation of Key Laboratory of Artificial Intelligence, Ministry of Education, P.R. China, and the Young Scholar Program (XSQD-202107001) from Beijing Institute of Technology.

{
    \small
    % \bibliographystyle{egbib}
    % \bibliography{egbib}

\begin{thebibliography}{10}

\bibitem{liu2023learnable}
S.~Liu, X.~Luo, K.~Fu, M.~Wang, and Z.~Song, ``A learnable self-supervised task for unsupervised domain adaptation on point cloud classification and segmentation,'' {\em Frontiers of Computer Science}, vol.~17, no.~6, p.~176708, 2023.

\bibitem{wu2023domain}
K.~Wu, F.~Jia, and Y.~Han, ``Domain-specific feature elimination: multi-source domain adaptation for image classification,'' {\em Frontiers of Computer Science}, vol.~17, no.~4, p.~174705, 2023.

\bibitem{zhu2023representation}
Y.~Zhu, X.~Wu, J.~Qiang, Y.~Yuan, and Y.~Li, ``Representation learning via an integrated autoencoder for unsupervised domain adaptation,'' {\em Frontiers of Computer Science}, vol.~17, no.~5, p.~175334, 2023.

\bibitem{Glasbey_Mardia_2002}
C.~A. Glasbey and K.~V. Mardia, ``A review of image-warping methods,'' {\em Journal of Applied Statistics}, vol.~25, p.~155–171, Jul 2002.

\bibitem{Zhou_2013_ICCV_Workshops}
{Zhou, Erjin and Fan, Haoqiang and Cao, Zhimin and Jiang, Yuning and Yin, Qi}, ``{Extensive Facial Landmark Localization with Coarse-to-Fine Convolutional Network Cascade},'' in {\em {Proceedings of the IEEE International Conference on Computer Vision (ICCV) Workshops}}, {June} {2013}.

\bibitem{Kowalski_2017_CVPR_Workshops}
M.~Kowalski, J.~Naruniec, and T.~Trzcinski, ``Deep alignment network: A convolutional neural network for robust face alignment,'' in {\em Proceedings of the IEEE Conference on Computer Vision and Pattern Recognition (CVPR) Workshops}, July 2017.

\bibitem{Li_2022_CVPR}
H.~Li, Z.~Guo, S.-M. Rhee, S.~Han, and J.-J. Han, ``Towards accurate facial landmark detection via cascaded transformers,'' in {\em Proceedings of the IEEE/CVF Conference on Computer Vision and Pattern Recognition (CVPR)}, pp.~4176--4185, June 2022.

\bibitem{vaswani2017attention}
A.~Vaswani, N.~Shazeer, N.~Parmar, J.~Uszkoreit, L.~Jones, A.~N. Gomez, {\L}.~Kaiser, and I.~Polosukhin, ``Attention is all you need,'' {\em Advances in neural information processing systems}, vol.~30, 2017.

\bibitem{Wang_2019_ICCV}
X.~Wang, L.~Bo, and L.~Fuxin, ``Adaptive wing loss for robust face alignment via heatmap regression,'' in {\em Proceedings of the IEEE/CVF International Conference on Computer Vision (ICCV)}, October 2019.

\bibitem{gatys2015neural}
L.~A. Gatys, A.~S. Ecker, and M.~Bethge, ``A neural algorithm of artistic style,'' {\em arXiv preprint arXiv:1508.06576}, 2015.

\bibitem{selim2016painting}
A.~Selim, M.~Elgharib, and L.~Doyle, ``Painting style transfer for head portraits using convolutional neural networks,'' {\em ACM Transactions on Graphics (ToG)}, vol.~35, no.~4, pp.~1--18, 2016.

\bibitem{wu2019landmark}
R.~Wu, X.~Gu, X.~Tao, X.~Shen, Y.-W. Tai, {\em et~al.}, ``Landmark assisted cyclegan for cartoon face generation,'' {\em arXiv preprint arXiv:1907.01424}, 2019.

\bibitem{shi2019warpgan}
Y.~Shi, D.~Deb, and A.~K. Jain, ``Warpgan: Automatic caricature generation,'' in {\em Proceedings of the IEEE/CVF conference on computer vision and pattern recognition}, pp.~10762--10771, 2019.

\bibitem{liu2021learning}
X.-C. Liu, Y.-L. Yang, and P.~Hall, ``Learning to warp for style transfer,'' in {\em Proceedings of the IEEE/CVF Conference on Computer Vision and Pattern Recognition}, pp.~3702--3711, 2021.

\bibitem{Zhu_2017_ICCV}
J.-Y. Zhu, T.~Park, P.~Isola, and A.~A. Efros, ``Unpaired image-to-image translation using cycle-consistent adversarial networks,'' in {\em Proceedings of the IEEE International Conference on Computer Vision (ICCV)}, Oct 2017.

\bibitem{pei2018multi}
Z.~Pei, Z.~Cao, M.~Long, and J.~Wang, ``Multi-adversarial domain adaptation,'' in {\em Thirty-second AAAI conference on artificial intelligence}, 2018.

\bibitem{long2015learning}
M.~Long, Y.~Cao, J.~Wang, and M.~Jordan, ``Learning transferable features with deep adaptation networks,'' in {\em International conference on machine learning}, pp.~97--105, PMLR, 2015.

\bibitem{long2016unsupervised}
M.~Long, H.~Zhu, J.~Wang, and M.~I. Jordan, ``Unsupervised domain adaptation with residual transfer networks,'' {\em Advances in neural information processing systems}, vol.~29, 2016.

\bibitem{ganin2015unsupervised}
Y.~Ganin and V.~Lempitsky, ``Unsupervised domain adaptation by backpropagation,'' in {\em International conference on machine learning}, pp.~1180--1189, PMLR, 2015.

\bibitem{zhang2018importance}
J.~Zhang, Z.~Ding, W.~Li, and P.~Ogunbona, ``Importance weighted adversarial nets for partial domain adaptation,'' in {\em Proceedings of the IEEE conference on computer vision and pattern recognition}, pp.~8156--8164, 2018.

\bibitem{cole2017synthesizing}
F.~Cole, D.~Belanger, D.~Krishnan, A.~Sarna, I.~Mosseri, and W.~T. Freeman, ``Synthesizing normalized faces from facial identity features,'' in {\em Proceedings of the IEEE conference on computer vision and pattern recognition}, pp.~3703--3712, 2017.

\bibitem{Kim_Kolkin_Salavon_Shakhnarovich_2020}
S.~Kim, N.~Kolkin, J.~Salavon, and G.~Shakhnarovich, ``Deformable style transfer,'' Mar 2020.

\bibitem{Cai_Guo_Peng_Zhang_2020}
H.~Cai, Y.~Guo, Z.~Peng, and J.~Zhang, ``Landmark detection and 3d face reconstruction for caricature using a nonlinear parametric model,'' Apr 2020.

\bibitem{Sagonas_Tzimiropoulos_Zafeiriou_Pantic_2014}
C.~Sagonas, G.~Tzimiropoulos, S.~Zafeiriou, and M.~Pantic, ``300 faces in-the-wild challenge: The first facial landmark localization challenge,'' in {\em 2013 IEEE International Conference on Computer Vision Workshops}, Mar 2014.

\bibitem{Belhumeur_Jacobs_Kriegman_Kumar_2011}
P.~N. Belhumeur, D.~W. Jacobs, D.~J. Kriegman, and N.~Kumar, ``Localizing parts of faces using a consensus of exemplars,'' in {\em CVPR 2011}, Aug 2011.

\bibitem{zhu2012face}
X.~Zhu and D.~Ramanan, ``Face detection, pose estimation, and landmark localization in the wild,'' in {\em 2012 IEEE conference on computer vision and pattern recognition}, pp.~2879--2886, IEEE, 2012.

\bibitem{Le_Brandt_Lin_Bourdev_Huang_2012}
V.~Le, J.~Brandt, Z.~Lin, L.~Bourdev, and T.~S. Huang, {\em Interactive Facial Feature Localization}, p.~679–692.
\newblock Sep 2012.

\bibitem{agbolade2020landmark}
O.~Agbolade, A.~Nazri, R.~Yaakob, A.~A. Abd~Ghani, and Y.~K. Cheah, ``Landmark-based homologous multi-point warping approach to 3d facial recognition using multiple datasets,'' {\em PeerJ Computer Science}, vol.~6, p.~e249, 2020.

\bibitem{dong2018supervision}
X.~Dong, S.-I. Yu, X.~Weng, S.-E. Wei, Y.~Yang, and Y.~Sheikh, ``Supervision-by-registration: An unsupervised approach to improve the precision of facial landmark detectors,'' in {\em Proceedings of the IEEE Conference on Computer Vision and Pattern Recognition}, pp.~360--368, 2018.

\bibitem{xia2022sparse}
J.~Xia, W.~Qu, W.~Huang, J.~Zhang, X.~Wang, and M.~Xu, ``Sparse local patch transformer for robust face alignment and landmarks inherent relation learning,'' in {\em Proceedings of the IEEE/CVF Conference on Computer Vision and Pattern Recognition}, pp.~4052--4061, 2022.

\bibitem{wang2020deep}
J.~Wang, K.~Sun, T.~Cheng, B.~Jiang, C.~Deng, Y.~Zhao, D.~Liu, Y.~Mu, M.~Tan, X.~Wang, {\em et~al.}, ``Deep high-resolution representation learning for visual recognition,'' {\em IEEE transactions on pattern analysis and machine intelligence}, vol.~43, no.~10, pp.~3349--3364, 2020.

\bibitem{kazemi2014one}
V.~Kazemi and J.~Sullivan, ``One millisecond face alignment with an ensemble of regression trees,'' in {\em Proceedings of the IEEE conference on computer vision and pattern recognition}, pp.~1867--1874, 2014.

\bibitem{taigman2014deepface}
Y.~Taigman, M.~Yang, M.~Ranzato, and L.~Wolf, ``Deepface: Closing the gap to human-level performance in face verification,'' in {\em Proceedings of the IEEE conference on computer vision and pattern recognition}, pp.~1701--1708, 2014.

\bibitem{masi2016pose}
I.~Masi, S.~Rawls, G.~Medioni, and P.~Natarajan, ``Pose-aware face recognition in the wild,'' in {\em Proceedings of the IEEE conference on computer vision and pattern recognition}, pp.~4838--4846, 2016.

\bibitem{liu2017sphereface}
W.~Liu, Y.~Wen, Z.~Yu, M.~Li, B.~Raj, and L.~Song, ``Sphereface: Deep hypersphere embedding for face recognition,'' in {\em Proceedings of the IEEE conference on computer vision and pattern recognition}, pp.~212--220, 2017.

\bibitem{yang2017neural}
J.~Yang, P.~Ren, D.~Zhang, D.~Chen, F.~Wen, H.~Li, and G.~Hua, ``Neural aggregation network for video face recognition,'' in {\em Proceedings of the IEEE conference on computer vision and pattern recognition}, pp.~4362--4371, 2017.

\bibitem{cao2018carigans}
K.~Cao, J.~Liao, and L.~Yuan, ``Carigans: Unpaired photo-to-caricature translation,'' {\em arXiv preprint arXiv:1811.00222}, 2018.

\bibitem{dou2017end}
P.~Dou, S.~K. Shah, and I.~A. Kakadiaris, ``End-to-end 3d face reconstruction with deep neural networks,'' in {\em proceedings of the IEEE conference on computer vision and pattern recognition}, pp.~5908--5917, 2017.

\bibitem{roth2015unconstrained}
J.~Roth, Y.~Tong, and X.~Liu, ``Unconstrained 3d face reconstruction,'' in {\em Proceedings of the IEEE conference on computer vision and pattern recognition}, pp.~2606--2615, 2015.

\bibitem{liu2016joint}
F.~Liu, D.~Zeng, Q.~Zhao, and X.~Liu, ``Joint face alignment and 3d face reconstruction,'' in {\em Computer Vision--ECCV 2016: 14th European Conference, Amsterdam, The Netherlands, October 11-14, 2016, Proceedings, Part V 14}, pp.~545--560, Springer, 2016.

\bibitem{lv2017deep}
J.~Lv, X.~Shao, J.~Xing, C.~Cheng, and X.~Zhou, ``A deep regression architecture with two-stage re-initialization for high performance facial landmark detection,'' in {\em Proceedings of the IEEE conference on computer vision and pattern recognition}, pp.~3317--3326, 2017.

\bibitem{karras2019style}
T.~Karras, S.~Laine, and T.~Aila, ``A style-based generator architecture for generative adversarial networks,'' in {\em Proceedings of the IEEE/CVF conference on computer vision and pattern recognition}, pp.~4401--4410, 2019.

\bibitem{gretton2006kernel}
A.~Gretton, K.~Borgwardt, M.~Rasch, B.~Sch{\"o}lkopf, and A.~Smola, ``A kernel method for the two-sample-problem,'' {\em Advances in neural information processing systems}, vol.~19, 2006.

\bibitem{sun2017correlation}
B.~Sun, J.~Feng, and K.~Saenko, ``Correlation alignment for unsupervised domain adaptation,'' {\em Domain adaptation in computer vision applications}, pp.~153--171, 2017.

\bibitem{kang2019contrastive}
G.~Kang, L.~Jiang, Y.~Yang, and A.~G. Hauptmann, ``Contrastive adaptation network for unsupervised domain adaptation,'' in {\em Proceedings of the IEEE/CVF conference on computer vision and pattern recognition}, pp.~4893--4902, 2019.

\bibitem{yaniv2019face}
J.~Yaniv, Y.~Newman, and A.~Shamir, ``The face of art: landmark detection and geometric style in portraits,'' {\em ACM Transactions on graphics (TOG)}, vol.~38, no.~4, pp.~1--15, 2019.

\bibitem{cao2014face}
X.~Cao, Y.~Wei, F.~Wen, and J.~Sun, ``Face alignment by explicit shape regression,'' {\em International journal of computer vision}, vol.~107, pp.~177--190, 2014.

\bibitem{li2019bidirectional}
Y.~Li, L.~Yuan, and N.~Vasconcelos, ``Bidirectional learning for domain adaptation of semantic segmentation,'' in {\em Proceedings of the IEEE/CVF Conference on Computer Vision and Pattern Recognition}, pp.~6936--6945, 2019.

\bibitem{zhou2023star}
Z.~Zhou, H.~Li, H.~Liu, N.~Wang, G.~Yu, and R.~Ji, ``Star loss: Reducing semantic ambiguity in facial landmark detection,'' in {\em Proceedings of the IEEE/CVF Conference on Computer Vision and Pattern Recognition}, pp.~15475--15484, 2023.

\bibitem{gong2020autotoon}
J.~Gong, Y.~Hold-Geoffroy, and J.~Lu, ``Autotoon: Automatic geometric warping for face cartoon generation,'' in {\em Proceedings of the IEEE/CVF winter conference on applications of computer vision}, pp.~360--369, 2020.

\bibitem{dosovitskiy2015flownet}
A.~Dosovitskiy, P.~Fischer, E.~Ilg, P.~Hausser, C.~Hazirbas, V.~Golkov, P.~Van Der~Smagt, D.~Cremers, and T.~Brox, ``Flownet: Learning optical flow with convolutional networks,'' in {\em Proceedings of the IEEE international conference on computer vision}, pp.~2758--2766, 2015.

\bibitem{teed2020raft}
Z.~Teed and J.~Deng, ``Raft: Recurrent all-pairs field transforms for optical flow,'' in {\em Computer Vision--ECCV 2020: 16th European Conference, Glasgow, UK, August 23--28, 2020, Proceedings, Part II 16}, pp.~402--419, Springer, 2020.

\bibitem{wu2018look}
W.~Wu, C.~Qian, S.~Yang, Q.~Wang, Y.~Cai, and Q.~Zhou, ``Look at boundary: A boundary-aware face alignment algorithm,'' in {\em Proceedings of the IEEE conference on computer vision and pattern recognition}, pp.~2129--2138, 2018.

\bibitem{kingma2014adam}
D.~P. Kingma and J.~Ba, ``Adam: A method for stochastic optimization,'' {\em arXiv preprint arXiv:1412.6980}, 2014.

\bibitem{messer1999xm2vtsdb}
K.~Messer, J.~Matas, J.~Kittler, J.~Luettin, G.~Maitre, {\em et~al.}, ``Xm2vtsdb: The extended m2vts database,'' in {\em Second international conference on audio and video-based biometric person authentication}, vol.~964, pp.~965--966, Citeseer, 1999.

\bibitem{Aliakbarian_2022_CVPR}
S.~Aliakbarian, P.~Cameron, F.~Bogo, A.~Fitzgibbon, and T.~J. Cashman, ``Flag: Flow-based 3d avatar generation from sparse observations,'' in {\em Proceedings of the IEEE/CVF Conference on Computer Vision and Pattern Recognition (CVPR)}, pp.~13253--13262, June 2022.

\bibitem{yang2020fda}
Y.~Yang and S.~Soatto, ``Fda: Fourier domain adaptation for semantic segmentation,'' in {\em Proceedings of the IEEE/CVF conference on computer vision and pattern recognition}, pp.~4085--4095, 2020.

\bibitem{tsai2018learning}
Y.-H. Tsai, W.-C. Hung, S.~Schulter, K.~Sohn, M.-H. Yang, and M.~Chandraker, ``Learning to adapt structured output space for semantic segmentation,'' in {\em Proceedings of the IEEE conference on computer vision and pattern recognition}, pp.~7472--7481, 2018.

\bibitem{cao2017realtime}
Z.~Cao, T.~Simon, S.-E. Wei, and Y.~Sheikh, ``Realtime multi-person 2d pose estimation using part affinity fields,'' in {\em Proceedings of the IEEE conference on computer vision and pattern recognition}, pp.~7291--7299, 2017.

\bibitem{prados2022shape}
A.~Prados-Torreblanca, J.~M. Buenaposada, and L.~Baumela, ``Shape preserving facial landmarks with graph attention networks,'' {\em arXiv preprint arXiv:2210.07233}, 2022.

\bibitem{wei2016convolutional}
S.-E. Wei, V.~Ramakrishna, T.~Kanade, and Y.~Sheikh, ``Convolutional pose machines,'' in {\em Proceedings of the IEEE conference on Computer Vision and Pattern Recognition}, pp.~4724--4732, 2016.

\bibitem{simonyan2014very}
K.~Simonyan and A.~Zisserman, ``Very deep convolutional networks for large-scale image recognition,'' {\em arXiv preprint arXiv:1409.1556}, 2014.

\bibitem{branwen2019danbooru2019}
G.~Branwen, ``Danbooru2019 portraits: A large-scale anime head illustration dataset,'' {\em Danbooru2019 portraits: A large-scale anime head illustration dataset}, 2019.

\end{thebibliography}

}

\clearpage
\newpage
\clearpage
\setcounter{page}{1}
\maketitlesupplementary

\section{Introduction}
This supplementary material provides the following information: 
% \cref{sec: algorithm} demonstrates a more comprehensive understanding of our learning scheme.
\cref{sec: implement} presents the implementation details in the experiments for the convenience of reproduction.
% \cref{sec: compare_sota} further validate the effectiveness of our proposed method by comparing with SOTA landmark detectors which may have the generalizability to unseen domains.
\cref{sec: supp_ablation} presents more ablation studies to further validate the effectiveness of our proposed framework.
\cref{sec: vis} provides sufficient visualization results as strong evidence for our method.
\cref{sec: future} indicates some failure cases predicted by our model, analyzes the possible limitations then points out the direction of future work.

\section{Training Details}\label{sec: implement}
We implement our landmark predictor $f_\theta$ as SLPT \citep{xia2022sparse} in the above experiments. 
Each input image is cropped and resized to $256 \times 256$, and the training set is augmented with various transformations such as random horizontal flipping, grayscale, occlusion, scaling, rotation, and translation. 
We select HRNetW180 \citep{wang2020deep} as the backbone model, with a feature map resolution of $64 \times 64$. 

To verify the impacts of different model architectures, we compare two different backbones: SBR \citep{dong2018supervision} and HRnet \citep{wang2020deep}.
For the SBR approach \citep{dong2018supervision}, we utilize CPM \citep{wei2016convolutional} as the detector, and VGG-16 \citep{simonyan2014very} networks to initial four convolutional layers for feature extraction and only three CPM stages are used for heatmap prediction. 
For the HRNet technique \citep{wang2020deep}, all faces are cropped based on their bounding boxes, centered using calculated formulas, and then resized to 256x256.
After that, we perform Data augmentation on images using in-plane rotation, scaling, and random flipping.

\section{More Ablation Studies}\label{sec: supp_ablation}

\subsection{Effect of different pose dataset}
\label{sec:Effect of different pose dataset}
Previously research employ warp methods such as AutoToon \citep{gong2020autotoon} and WarpGAN \citep{shi2019warpgan} for facial manipulation, as well as common flow prediction methods like FlowNet \citep{dosovitskiy2015flownet} and RAFT \citep{teed2020raft}. 
These methods either require a one-to-one correspondence between input images or assume minimal deformation between two images. 
Consequently, for facial images, the positioning of the face also influences the results. 
However, establishing a one-to-one correspondence between the 300W dataset and the CariFace dataset is challenging and time-consuming. 
To address this issue, we consider categorizing the datasets into three classes: frontal faces, faces turned right, and faces turned left, each comprising 1000 images. 
Subsequently, separate training is conducted for each category, and the results can be observed in Table \ref{ablation_dataset}.

% \begin{table}[h]
% \caption{Ablation study on the usage of the dataset}\label{ablation_dataset}
% \centering
% \begin{tabular*}{0.4\textwidth}{@{\extracolsep\fill}lcc}
% \toprule%
%   \multicolumn{2}{@{}c@{}}{Direction} & \multirow{2}{*}{Results}\\\cmidrule{1-2}%
%  300W & Cariface \\
% \midrule
% ALL & ALL &  9.374 \\
% Right & Frontal & 9.519 \\
% Left & Frontal &  9.518\\
% Frontal & Frontal & 9.518 \\
% Right & ALL & 9.376\\
% Left & ALL & 9.372 \\
% Frontal & ALL & \color{red}{9.370} \\
% \bottomrule
% \end{tabular*}
% \end{table}

% \begin{table*}[h]
% \caption{Ablation study on the usage of the dataset.}\label{ablation_dataset}
% \centering
% \begin{tabular}{cc|c|c|c|c|c|c|c}
% \toprule%
% \multicolumn{2}{c|}{\multirow{2}{*}{Direction}} & \multicolumn{4}{c|}{300W} & \multicolumn{2}{c|}{CariFace} & \multirow{2}{*}{Result} \\
% \cline{3-6} \cline{7-8}
% & & ALL & Frontal & Left & Right & ALL & Frontal\\
% \midrule
% \multicolumn{2}{c|}{\multirow{7}{*}{Usage}} & \checkmark &  & & & \checkmark & & 9.374 \\
% && &  & & \checkmark & & \checkmark & 9.519\\
% && &  &\checkmark & & & \checkmark & 9.518\\
% && & \checkmark & & & & \checkmark & 9.518\\
% && & & & \checkmark  & \checkmark & & 9.376\\
% && & & \checkmark  & & \checkmark & & 9.372\\
% &&  & \checkmark& & & \checkmark & &  \color{red}{9.370}\\
% \bottomrule
% \end{tabular}
% \end{table*}
\begin{figure*}[t]
    \centering
    \begin{subfigure}[b]{1\textwidth}
        \centering
        \includegraphics[height=3.0cm]{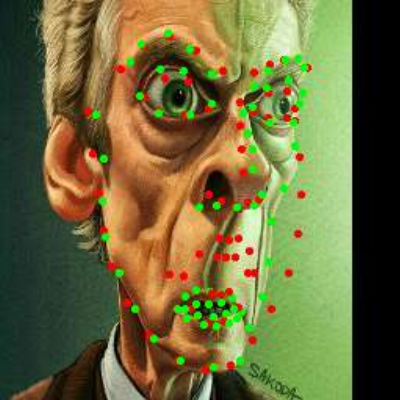}
        \includegraphics[height=3.0cm]{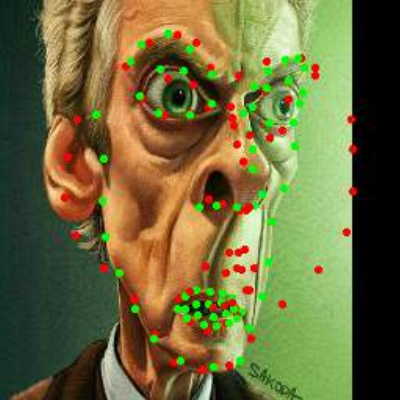}
        \includegraphics[height=3.0cm]{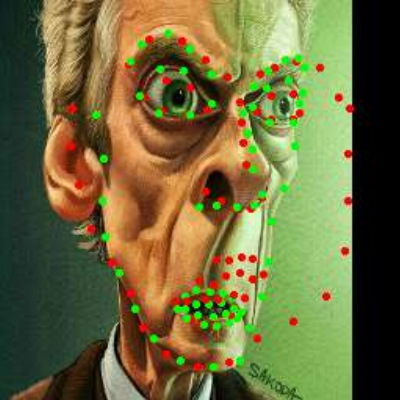}
        \includegraphics[height=3.0cm]{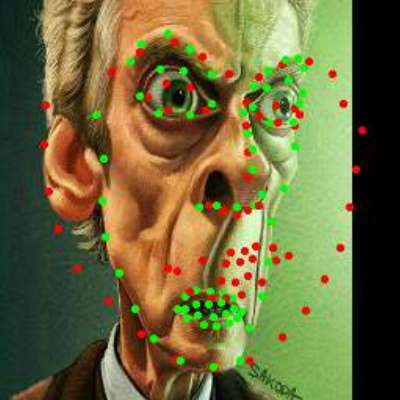}
        \includegraphics[height=3.0cm]{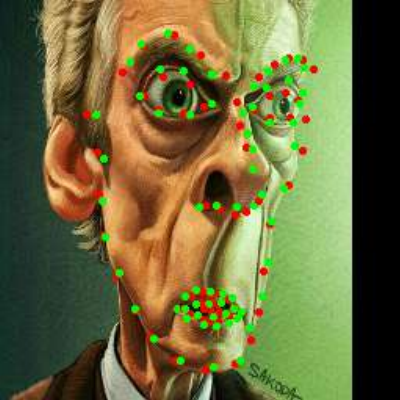}\\
       \includegraphics[width=3.0cm]{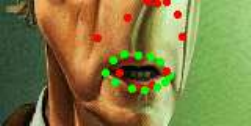}
        \includegraphics[width=3.0cm]{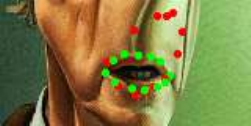}
        \includegraphics[width=3.0cm]{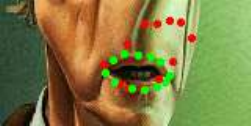}
        \includegraphics[width=3.0cm]{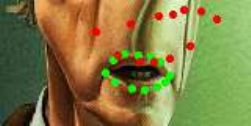}
        \includegraphics[width=3.0cm]{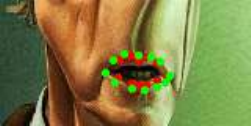}\\
        \small{SLPT (Source-only)\hspace{1.1cm} RevGrad\hspace{2cm}CycleGAN\hspace{2cm} BDL\hspace{2.2cm} Ours\hspace{1cm}}
        \subcaption{Case 1}
    \end{subfigure}
    \begin{subfigure}[b]{1\textwidth}
        \centering
       \includegraphics[height=3.0cm]{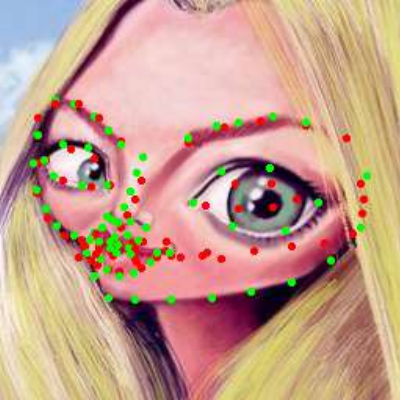}
        \includegraphics[height=3.0cm]{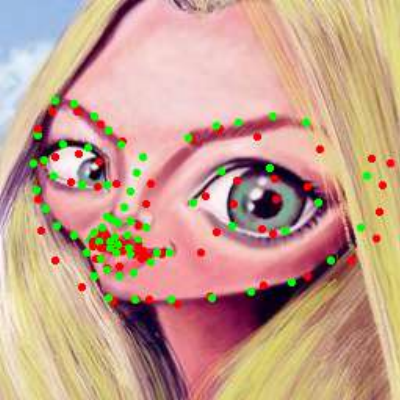}
        \includegraphics[height=3.0cm]{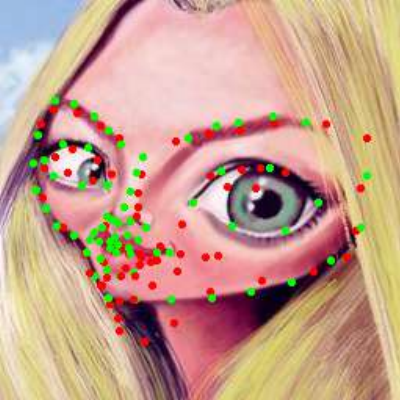}
        \includegraphics[height=3.0cm]{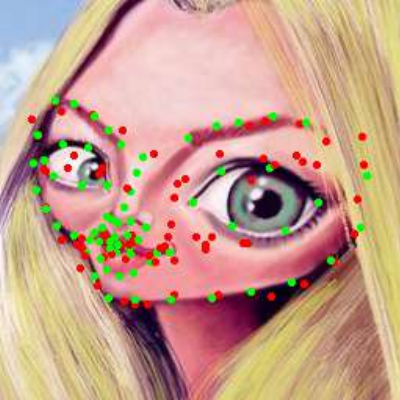}
        \includegraphics[height=3.0cm]{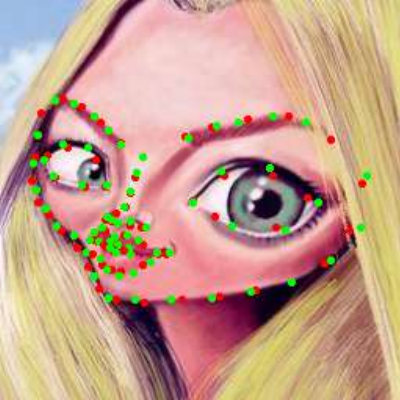}\\
       \includegraphics[width=3.0cm]{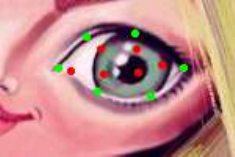}
        \includegraphics[width=3.0cm]{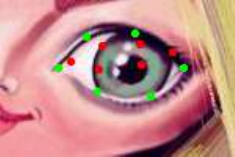}
        \includegraphics[width=3.0cm]{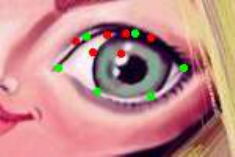}
        \includegraphics[width=3.0cm]{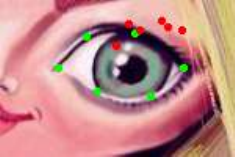}
        \includegraphics[width=3.0cm]{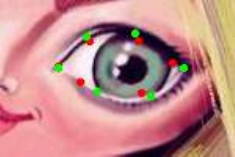}\\
        \small{SLPT (Source-only)\hspace{1.1cm} RevGrad\hspace{2cm}CycleGAN\hspace{2cm} BDL\hspace{2.2cm} Ours\hspace{1cm}}
        \subcaption{Case 2}
    \end{subfigure}
    \caption{ Visual comparisons for various methods in the two DA(300W$\rightarrow$CariFace) settings.}
    \label{fig:cmp_300W}
\end{figure*}

\begin{table}[h]
\caption{Ablation study on the usage of the dataset.}\label{ablation_dataset}
\centering
\begin{tabular*}{0.48\textwidth}{c|c|cccc}
% \toprule%
\hline\hline
\multicolumn{2}{c|}{\multirow{2}{*}{\textbf{Settings}}} & \multicolumn{4}{c}{300W}\\
\cline{3-6}
\multicolumn{2}{c|}{} & ALL & Frontal & Left & Right\\
% \midrule
\hline
\multirow{2}{*}{CariFace} & ALL & 7.831 & \color{red}{7.695} & 7.879 & 8.080 \\
& Frontal &8.466 & 9.077 & 8.768 &7.771  \\
% \bottomrule
\hline\hline
\end{tabular*}
\end{table}

From these results, we can observe variations in the outcomes when training with datasets containing different poses. 
Additionally, it is evident that training with all available datasets does not necessarily guarantee improved performance. 
Notably, the best results are achieved when utilizing the frontal 300W dataset in conjunction with all cartoon datasets. 
This could be attributed to the enhanced flexibility and effectiveness of warping processes when performed from a frontal perspective. 
That's why we choose this setting for our experiments.
This innovative strategy not only improves the overall accuracy and reliability of facial landmark prediction but also simplifies the training process.
% Furthermore, we observe a noteworthy trend in our experiments, where the model trained on a combination of the 300W dataset and corresponding datasets showcases the most favorable performance. 
% This outcome can be attributed to the augmented training data resulting from the amalgamation of these diverse datasets. 

\section{More Visualization}\label{sec: vis}
We present more samples to show landmark prediction results of our method under different styles and textures in the CariFace dataset, such as different facial expressions, various head poses, illumination, etc.

\begin{figure*}[t]
    \centering
    \begin{subfigure}[b]{1\textwidth}
        \centering
       \includegraphics[height=3.0cm]{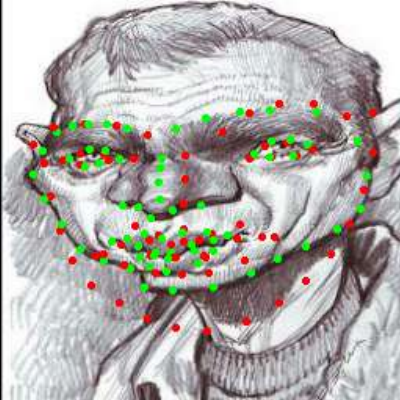}
        \includegraphics[height=3.0cm]{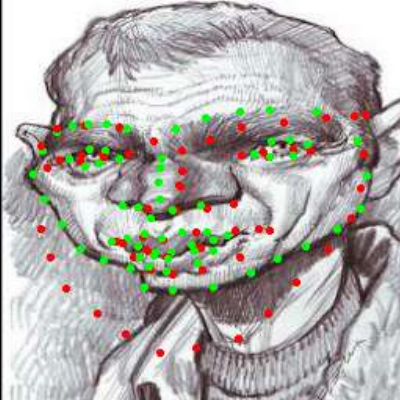}
        \includegraphics[height=3.0cm]{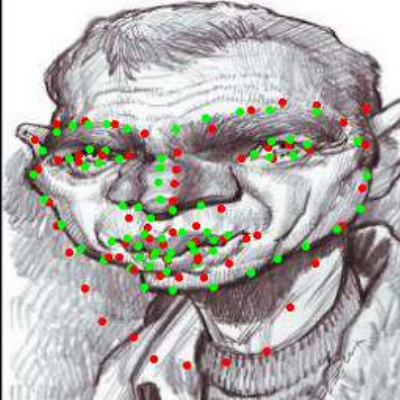}
        \includegraphics[height=3.0cm]{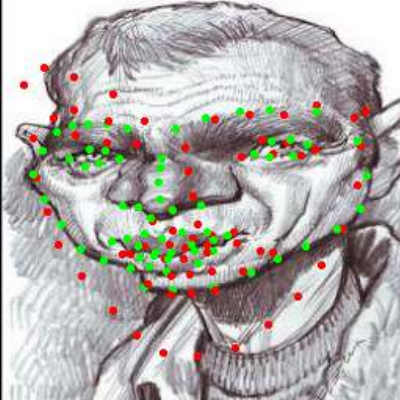}
        \includegraphics[height=3.0cm]{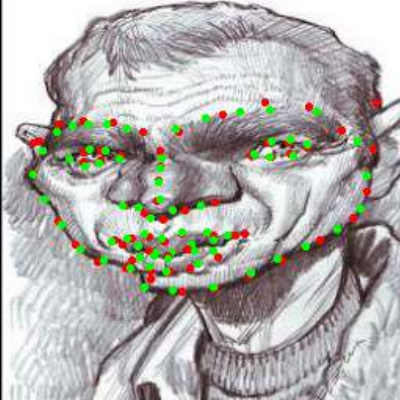}\\
       \includegraphics[width=3.0cm]{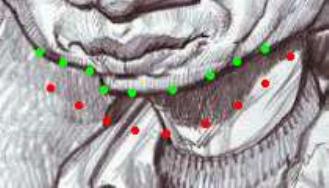}
        \includegraphics[width=3.0cm]{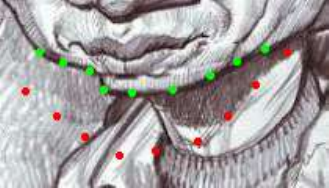}
        \includegraphics[width=3.0cm]{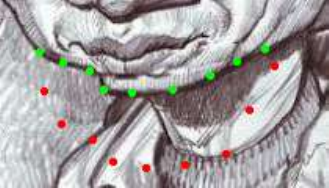}
        \includegraphics[width=3.0cm]{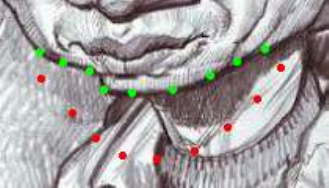}
        \includegraphics[width=3.0cm]{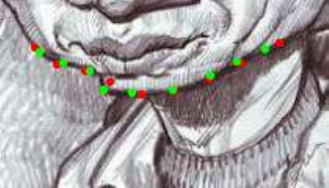}\\
        \small{SLPT (Source-only)\hspace{1.1cm} RevGrad\hspace{2cm}CycleGAN\hspace{2cm} BDL\hspace{2.2cm} Ours\hspace{1cm}}
        \subcaption{Case 3}
    \end{subfigure}
    \begin{subfigure}[b]{1\textwidth}
        \centering
       \includegraphics[height=3.0cm]{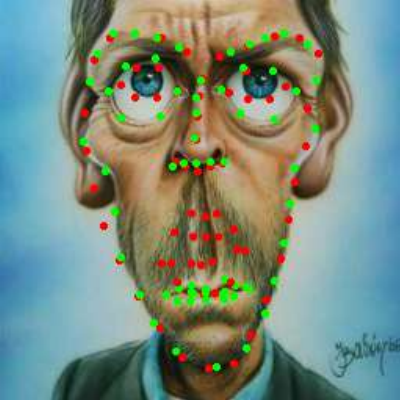}
        \includegraphics[height=3.0cm]{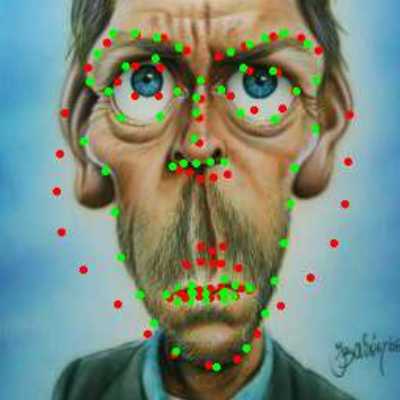}
        \includegraphics[height=3.0cm]{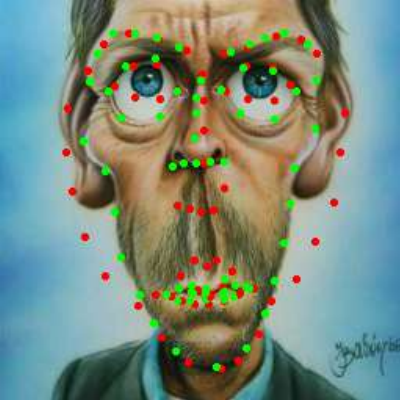}
        \includegraphics[height=3.0cm]{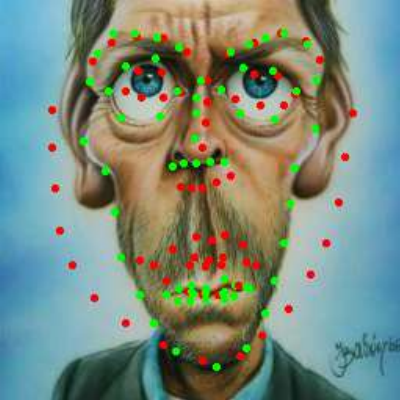}
        \includegraphics[height=3.0cm]{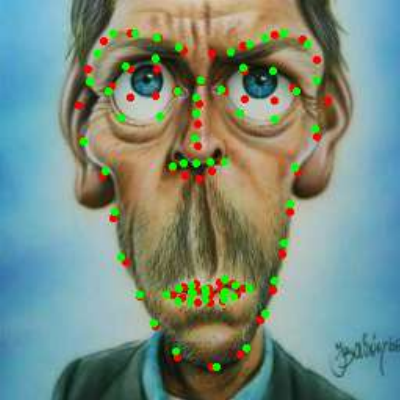}\\
       \includegraphics[width=3.0cm]{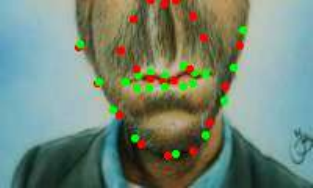}
        \includegraphics[width=3.0cm]{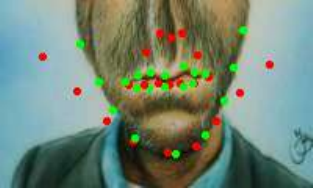}
        \includegraphics[width=3.0cm]{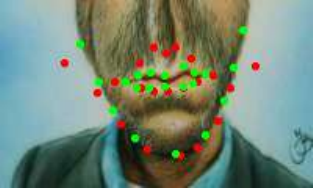}
        \includegraphics[width=3.0cm]{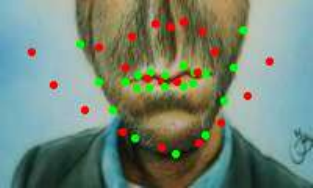}
        \includegraphics[width=3.0cm]{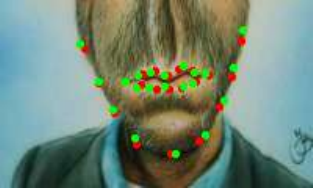}\\
        \small{SLPT (Source-only)\hspace{1.1cm} RevGrad\hspace{2cm}CycleGAN\hspace{2cm} BDL\hspace{2.2cm} Ours\hspace{1cm}}
        \subcaption{Case 4}
    \end{subfigure}
    \caption{ Visual comparisons for various methods in the two DA(300W$\rightarrow$CariFace) settings.}
    \label{fig:cmp_300W-Cariface}
\end{figure*}
\cref{fig:cmp_300W} and ~\cref{fig:cmp_300W-Cariface} demonstrate the effectiveness of our method in accurately predicting facial landmarks across various scenarios, including instances with exaggerated facial features. 
% Based on case 1, we can see that our method is able to more accurately predict where the mouth is when the distance between the nose and the mouth is larger.
In Case 1, our method excels at determining the mouth's position when the distance between the nose and the mouth is significantly larger.
Case 2 highlights our method's ability to precisely predict the eye edges when they are notably larger than other facial components.
Furthermore, Case 3 and 4 showcase our method's capability to accurately estimate facial contours when the face is compressed in both vertical and horizontal directions.

\section{Limitations and Discussion}\label{sec: future}
Based on our extensive experiments, our proposed method has achieved impressive results in unsupervised cartoon face landmark detection.
Notably, our model exhibits robust performance even when applied to previously unseen domains, surpassing some supervised approaches in certain cases. 

\begin{figure}
    \centering
    \includegraphics[width=0.48\textwidth]{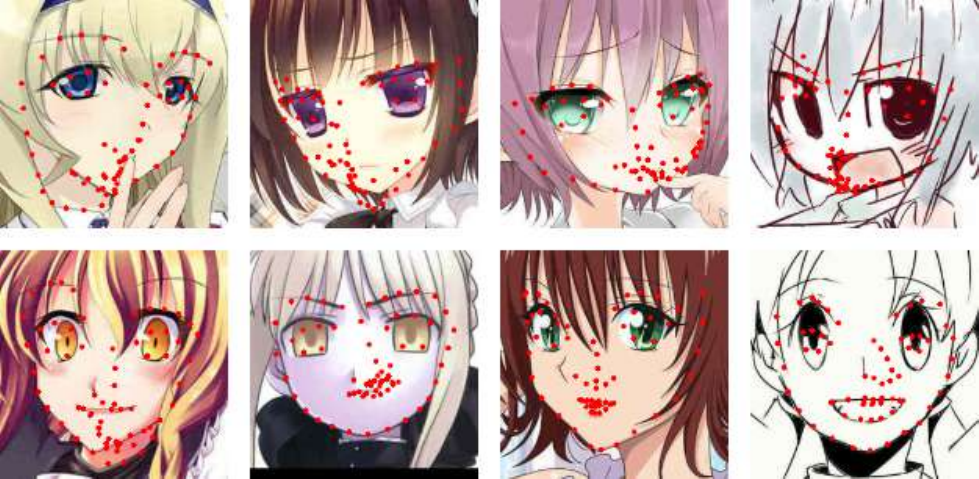}
    \caption{Visualizations of some typical failures. Red dots represent our predictions.}
    \label{fig: our_failure}
\end{figure}

Yet, there is still progress for improvements particularly in challenging situations, such as severe occlusion, blurring, and extremely stylized facial contours, as illustrated in \cref{fig: our_failure}.
To address these limitations, we have identified two specific issues: 
1) our model may struggle to predict facial contours when there is uncertainty in the face boundary.
2) when applied to more challenging scenarios, such as anime dataset \citep{branwen2019danbooru2019}, our model encounters difficulty in adapting to the domain which is characterized by distinctive features like small noses, mouths, and larger eyes.

For these weaknesses, the most possible solution is to construct a more robust constraint between the warped faces and cartoon faces for better prediction.
We leave it in the future work.

\end{document}